%% file: neurips_2024.tex
\documentclass{article}

\usepackage[preprint]{neurips_2024}

\usepackage[utf8]{inputenc} 
\usepackage[T1]{fontenc}    
\usepackage{hyperref}       
\usepackage{url}            
\usepackage{booktabs}       
\usepackage{amsfonts}       
\usepackage{nicefrac}       
\usepackage{microtype}      
\usepackage{xcolor}         
\usepackage{graphicx}
\usepackage{hyperref}
\usepackage{amsmath}
\DeclareMathOperator*{\argmax}{arg\,max}

\title{Defining error accumulation in \\ ML atmospheric simulators}

\author{%
  Raghul Parthipan \\
  University of Cambridge\\
  British Antarctic Survey \\
  \texttt{rp542@cam.ac.uk} \\
  \And
  Mohit Anand \\
  Helmholtz Centre for Environmental Research - UFZ \\
  Technische Universit\"at Dresden \\
  \texttt{mohit.anand@ufz.de} \\
  \AND
  Hannah M. Christensen \\
  University of Oxford \\
    \texttt{hannah.christensen@physics.ox.ac.uk} \\
  \And
  J. Scott Hosking \\
  The Alan Turing Institute \\
  British Antarctic Survey \\
  \texttt{jask@bas.ac.uk} \\
  \And
  Damon J. Wischik \\
  University of Cambridge \\
  \texttt{djw1005@cam.ac.uk} \\
}

\begin{document}

\maketitle

\begin{abstract}
  Machine learning (ML) has recently shown significant promise in modelling atmospheric systems, such as the weather. Many of these ML models are autoregressive, and error accumulation in their forecasts is a key problem. However, there is no clear definition of what `error accumulation' actually entails. In this paper, we propose a definition and an associated metric to measure it. Our definition distinguishes between errors which are due to model deficiencies, which we may hope to fix, and those due to the intrinsic properties of atmospheric systems (chaos, unobserved variables), which are not fixable. We illustrate the usefulness of this definition by proposing a simple regularization loss penalty inspired by it. This approach shows performance improvements (according to RMSE and spread/skill) in a selection of atmospheric systems, including the real-world weather prediction task.
\end{abstract}

\input{chapter1/document}

\input{chapter2/document}

\input{chapter3/document}

\input{chapter4/document}

\input{chapter5/document}

{
\small
\bibliographystyle{plainnat}
\bibliography{bibliography.bib}
}


\appendix

\input{appendix}


\newpage

\end{document}

%% file: chapter1/document.tex
\section{Introduction}

Error accumulation is a well-known problem when modelling dynamical systems. The issue has been highlighted by the burgeoning work doing Numerical Weather Prediction (NWP) with machine learning (ML) \citep{bi2023accurate,chen2023fengwu,chen2023fuxi,lam2022graphcast,pathak2022fourcastnet,nguyen2023scaling}. It is also evident in studies on simpler atmospheric systems such as the Lorenz 63 \citep{balogh2021toy,dubois2020data,pyle2021domain} and the Lorenz 96 \citep{gan_hannah,parthipan2023using,pyle2021domain}, as well as in work modelling fluids \citep{sanchez2020learning}. 

Despite its prevalence, there is no standard definition for what error accumulation actually is. Informally, it is understood as the errors that arise when an autoregressive  model is iterated forward in time to create a forecast. Small errors are made at each simulation step, and these can compound over time. The more extreme consequences of this --- a simulator exploding and returning NaNs, or becoming unstable --- are easy to diagnose. But more subtle forms, where the simulator remains stable but no longer generates trajectories that resemble the truth, are harder to detect. Typically, a combination of metrics such as root-mean-squared-error (RMSE), skill/spread, and continuous ranked probability score (CRPS) must be used to assess this.

In this work, we seek to clarify the types of errors that should be included in a definition of error accumulation (Section \ref{section:problem_formulation}), and then propose an appropriate definition along with a corresponding metric to measure this (Section \ref{section:def}). Our definition differentiates between errors that arise from the iterative process that are: i) due to ML model deficiencies (e.g. explosive simulations) and therefore types of errors we may hope to fix, and ii) those that arise from intrinsic properties of atmospheric systems, such as being chaotic and often having unobserved variables, which we may not hope to remedy. Our metric assesses this by evaluating model performance against a reference model that is immune to errors from iterative rollouts. Both the model to be assessed, and the reference model are for the same system, so both are limited by chaos/unobserved variables. This controls for the unfixable errors, allowing the metric to highlight model-deficiency errors.

We believe that a clear definition of error accumulation, which highlights errors due to model deficiencies, can aid in evaluating models, debugging models and guiding future model development. Additionally, a clear definition may suggest potential remedies. For example, in Section \ref{section:solution}, we propose a simple regularization loss penalty inspired by our definition. We test this on the Lorenz 63 (a simple chaotic atmospheric simulator), Lorenz 96 (a more complex atmospheric simulator) and the real-world weather prediction task (using ERA5 data), observing performance improvements. We do not claim this is the definitive remedy for error accumulation, but use this to illustrate how our definition can help identify solutions.

%% file: chapter2/document.tex
\section{Background}

\paragraph{Defining error accumulation in dynamical systems.} There is no widely-used definition for error accumulation. ML-based weather prediction (MLWP) work \citep{bi2023accurate,chen2023fuxi,nguyen2023scaling,lam2022graphcast}, ML parameterization work \citep{brenowitz2018prognostic,brenowitz2019spatially} and dynamical systems work \citep{liu2020hierarchical,sanchez2020learning} all mention error accumulation, typically describing it informally as a progressive increase in forecast errors with each iteration of an autoregressive forecast. \citet{balogh2021toy} provide a `stability criterion' to determine if a trajectory remains stable by checking if it stays within seven times the range of the training data, acknowledging the  simplicity of this method in capturing explosive trajectories. Identifying more subtle forms of error accumulation requires using a mixture of metrics such as RMSE (Section \ref{appendix:rmse}), spread/skill (Section \ref{appendix:spread_skill}) and CRPS (Section \ref{appendix:crps}).


\paragraph{Addressing error accumulation.}

A common belief is that error accumulation results from a mismatch between the typical way of training autoregressive models (by maximizing likelihood), where true past states are conditioned on, and their use at simulation time, where model-generated states are conditioned on instead. This discrepancy, known as `exposure bias' in the language domain \citep{bengio_scheduled_sampling,lamb2016professor,leblond2017searnn,ranzato2015sequence,schmidt2019generalization}, has prompted the development of rollout training strategies. These involve generating model trajectories and using a loss function to penalize the difference between generated and true trajectories. Initially introduced as Scheduled Sampling in the language domain \citep{bengio_scheduled_sampling}, this approach has been adapted for MLWP such as GraphCast \citep{lam2022graphcast}, Stormer \citep{nguyen2023scaling} and FuXi \citep{chen2023fuxi} as well as for ML parameterization work \citep{brenowitz2018prognostic,brenowitz2019spatially}. However, the need to sample  trajectories during training significantly increases compute and memory requirements. \citep{chen2023fengwu} use a replay buffer strategy to alleviate it. For models based on techniques such as diffusion, it is not clear how rollout training would be used.

\citet{schmidt2019generalization} argues that the training by maximizing likelihood is not the issue; rather, this goal is the natural consequence of fitting probability models to data. Instead, poor test-time performance indicates a lack of model generalization, suggesting better regularization is needed. Rollout training can be seen as a form of regularization through data augmentation, using synthetic (generated) data during training. \citet{sanchez2020learning} and \citet{ho2022cascaded} are more explicit about this by adding Gaussian noise to their training inputs. \citet{balogh2021toy} stabilize ML simulations by training on data from regions of the dynamical system away from the attractor.  Similarly, \citet{rasp2020} treats high-resolution simulations as ground truth, running them in parallel with the ML model. Classic L2 regularization has been used to improve stability in ML parameterization \citep{bretherton2022correcting}, and regularization has also been applied through imposing physical constraints \citep{brenowitz2019spatially,yuval2021}.

\citet{liu2020hierarchical} and \citet{bi2023accurate} move away from standard autoregressive models to hierarchical models that forecast at different time horizons, minimizing the number of times that model-generated data is fed back into the model. This reduces error accumulation, as there are fewer opportunities for errors created by a model to propagate; however, this approach can lead to temporally inconsistent forecasts.

%% file: chapter3/document.tex
\section{Problem formulation: capturing error accumulation}
\label{section:problem_formulation}

In this section, we describe what a useful error accumulation definition and associated metric should capture. We also discuss how existing metrics fall short in certain aspects. 

\begin{figure}[h!]
  \centering
  \hspace*{-1.1cm}
  \includegraphics[clip, trim={0.1cm 0cm 0cm 0cm},clip,height=3.8cm]{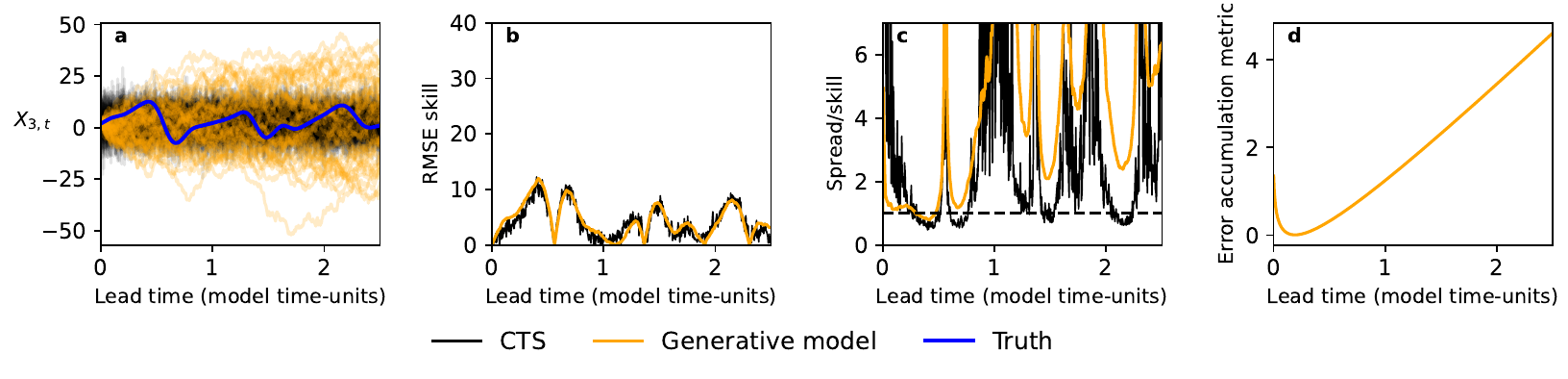}
  \caption{Illustration of a form of explosive error accumulation in the Lorenz 96. (a) Simulations from the random walk generative model (equation \ref{eq:random_walk}) and the CTS  (equation \ref{eq:cts_l96}) for a specific initial condition, with the ground truth (blue). (b)  RMSE, (c) spread/skill, and (d) our error accumulation metric. The spread/skill is erratic as the RMSE is often near zero, leading to large  spread/skill values. Despite this, the spread/skill of the generative model increases over time (visible in the increasing spread of trajectories in (a)), unlike the CTS. The generative model's poor behaviour (explosive trajectories) is most evident in (d).} 
  \label{fig:err_acc_explosion}
\end{figure}

\paragraph{Capturing explosive forms.} This includes obvious cases where the ensemble mean of the forecast grows indefinitely, as well as subtle cases where individual samples explode even if the ensemble mean remains stable. We illustrate this with an example from the Lorenz 96, using a generative model that will almost surely explode --- the random walk:
\begin{equation}
    X_{k,t+1} = X_{k,t} + Z_{k,t}
    \label{eq:random_walk}
\end{equation}
where $k$ is a spatial coordinate and $Z_{k,t} \sim N(0,1)$. Figure \ref{fig:err_acc_explosion}a shows trajectories for $X_{3,t}$ given an initial condition. As lead time increases, more trajectories from the generative model start to explode. However, the RMSE in Figure \ref{fig:err_acc_explosion}b fails to capture this because the mean of the generative model trajectories remains close to the truth (blue line), so the explosion of particular samples is not picked up. Only when combined with the spread/skill (Figure \ref{fig:err_acc_explosion}c) does this become clearer with how the spread/skill increases over time, reflecting the diverging trajectories.

\begin{figure}[h!]
  \centering
  \hspace*{-1.1cm}
  \includegraphics[clip, trim={0.1cm 0cm 0cm 0cm},clip,height=3.8cm]{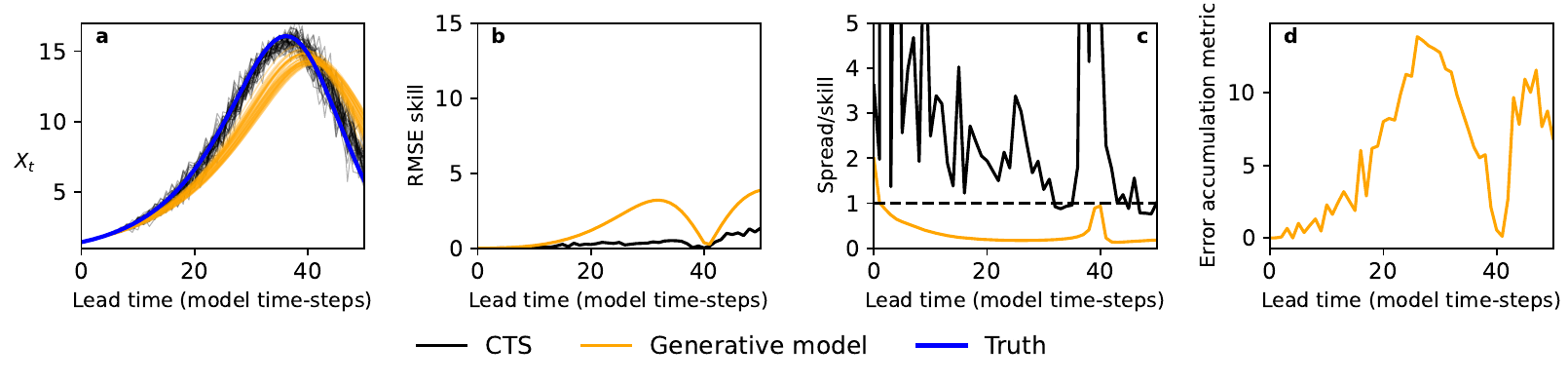}
  \caption{Illustration of a non-explosive form of error accumulation in the Lorenz 63. (a) Simulations from an iterative generative model and a CTS for a specific initial condition, with the ground truth (blue). The generative model trajectories fail to cover the truth, especially for lead times from 20 to 30. This issue is not apparent from the relatively small RMSE in (b) and can only inferred  from the under-dispersion in the spread/skill in (c). (d) Our error accumulation metric captures the generative model's failure to capture the truth, whilst the CTS does, suggesting the generative model's errors are due to model deficiencies as opposed to other factors.} 
  \label{fig:err_acc_non_explosion}
\end{figure}

\paragraph{Capturing non-explosive forms.}  Figure \ref{fig:err_acc_non_explosion}a shows an example from the Lorenz 63 where the generative model's trajectories fail to cover the true trajectory between lead times of 20 to 30. We know it is possible to capture the true trajectory though, as evidenced by our CTS (black), which suggests the generative model's failure is due to model deficiency. This failure is not apparent from the RMSE (Figure \ref{fig:err_acc_non_explosion}b) which remains relatively low. Only the under-dispersion in the spread/skill (Figure \ref{fig:err_acc_non_explosion}c) hints at it. 

\begin{figure}[h!]
  \centering
  \hspace*{-1.1cm}
  \includegraphics[clip, trim={0.1cm 0cm 0cm 0cm},clip,height=3.8cm]{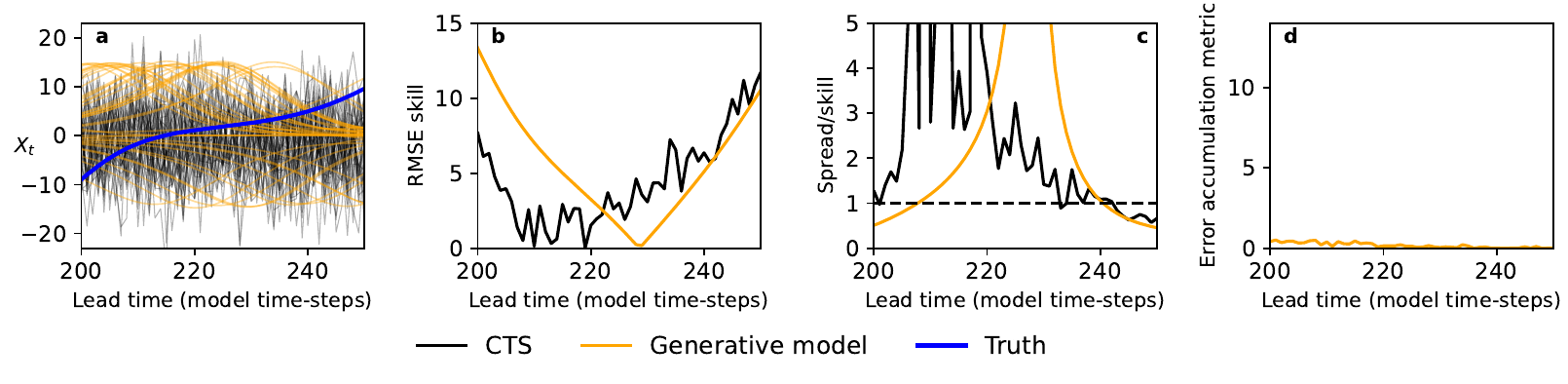}
  \caption{Example of predictability limits being reached due to STIC in the Lorenz 63. (a) Simulations from Figure \ref{fig:err_acc_non_explosion} extended to further lead times. The predictability horizon having been reached is not immediately obvious from (b) RMSE nor (c) spread/skill. (d) our error accumulation metric remains small, indicating that the generative model performs similarly to the CTS, suggesting the remaining errors are due to factors (STIC) separate to model deficiency.} 
  \label{fig:err_acc_stic_explosion}
\end{figure}

\paragraph{Not capturing errors due to STIC.}  Errors from Sensitivity To Initial Conditions (STIC) are due to properties of the underlying system, not failings of the ML model. Therefore, we wish to exclude this from our definition. In chaotic systems, STIC imposes a predictability limit, beyond which model states can diverge widely \citep{lorenz1969predictability,lorenz1996predictability}. For example, NWP models show increasing forecast errors with time, reaching their predictability limit at about two weeks \citep{judt2018insights,zhang2019predictability}. Beyond this horizon, forecasts stabilize around climatological distributions without accurately predicting the specific future evolution.  This error growth is not correctable. Figure \ref{fig:err_acc_stic_explosion}a illustrates this with the generative model reaching the predictablility limit of the chaotic Lorenz 63.  Although this is expected behaviour, it is not apparent in the RMSE (\ref{fig:err_acc_stic_explosion}b) which varies considerably.

\paragraph{Not capturing errors due to unobserved variables.} The spatio-temporal resolution of input variables impacts how well ML models can predict a system's evolution. Higher resolution has driven improvements in NWP forecast skill over the past years \citep{bauer2015quiet}. This trend is also reflected in MLWP, where finer resolution forecasts \citep[e.g.][]{bi2023accurate,lam2022graphcast,nguyen2023scaling} typically yield better results compared to coarser ones \citep[e.g.][]{clare2021combining,rasp2020weatherbench,rasp2021data}. Thus, error reduction is an expected outcome of higher-resolution data, indicating a natural skill limit imposed by the data resolution rather than the ML model itself.

\paragraph{Existing metrics lack a reference point for highlighting error accumulation.} Existing metrics often have to be combined and interpreted together to assess fixable error accumulation. For example, An RMSE of zero can mask explosive trajectories (Figure \ref{fig:err_acc_explosion}). Even when combined, these metrics lack clarity on attainable scores --- RMSEs of zero may not be achievable due to STIC/unobserved variables. This issue would be mitigated with a clear reference model against which to compare RMSE, spread/skill and other metrics. However, a clear reference model is also lacking as the gold-standard for weather comparisons, the ECMWF's ENS and HRES, are often surpassed by existing ML models \citep{bi2023accurate,chen2023fuxi,lam2022graphcast,price2023gencast} so cannot highlight performance gaps which may still exist.

\section{Defining error accumulation}
\label{section:def}

We propose a definition that measures the \emph{fixable types} of errors arising from the iterative process, and provides a \emph{clear reference point} for how small we can hope to make these errors. 

We define error accumulation over a period of time, $T$, as the growth of $\delta(t)$ over $T$, where
\begin{equation}
    \delta(t) =  \mathbb{E}_{x_{1:c} \sim \mathrm{data}} \,  \mathrm{KL}\big(p_{\mathrm{gen}}(x_{t+c}|x_{1:c}) \, || \, p_{\mathrm{truth}}(x_{t+c}|x_{1:c})\big)
    \label{eq:err_acc_kl}
\end{equation}
The expectation is taken over possible  $x$ sequences drawn from the data distribution, with $x_{1:c}$ as context data (e.g. initial conditions). Here, $p_{\mathrm{gen}}$ is the generative model being assessed, and $p_{\mathrm{truth}}$ is the truth model. 

$p_{\mathrm{truth}}$ serves as a clear reference because it encapsulates system properties like chaos and unobserved variables. Therefore, any discrepancies between $p_{\mathrm{truth}}$ and $p_{\mathrm{gen}}$ must be attributed to deficiencies in $p_{\mathrm{gen}}$. 

Since $p_{\mathrm{truth}}$ is typically inaccessible, we approximate it with a continuous forecasting model (CTS), $p_{\mathrm{cts}}$, such that
\begin{equation}
       \delta(t) \approx  \mathbb{E}_{x_{1:c}\sim \mathrm{data}} \,  \mathrm{KL}\big(p_{\mathrm{gen}}(x_{t+c}|x_{1:c}) \, || \, p_{\mathrm{cts}}(x_{t+c}|x_{1:c})\big)
    \label{eq:err_acc_kl_approx} 
\end{equation}
A CTS models $x_{t+c}$ directly from  $x_{1:c}$ and $t$ without feeding back its predictions. This makes the CTS a useful approximation for $p_\mathrm{truth}$ because even though we train it, it is immune to error accumulation. Thus, the CTS becomes our reference model. It is also constrained by system properties, meaning any improvement in $p_\mathrm{cts}$ over $p_\mathrm{gen}$ highlights deficiencies in $p_\mathrm{gen}$ that can potentially be fixed, rather than issues inherent to the data or system. We henceforth refer to equation \ref{eq:err_acc_kl_approx} as our error accumulation metric.

\begin{figure}[h!]
  \centering
  \includegraphics[clip, trim={3cm 4.5cm 2cm 2cm},clip,height=8cm]{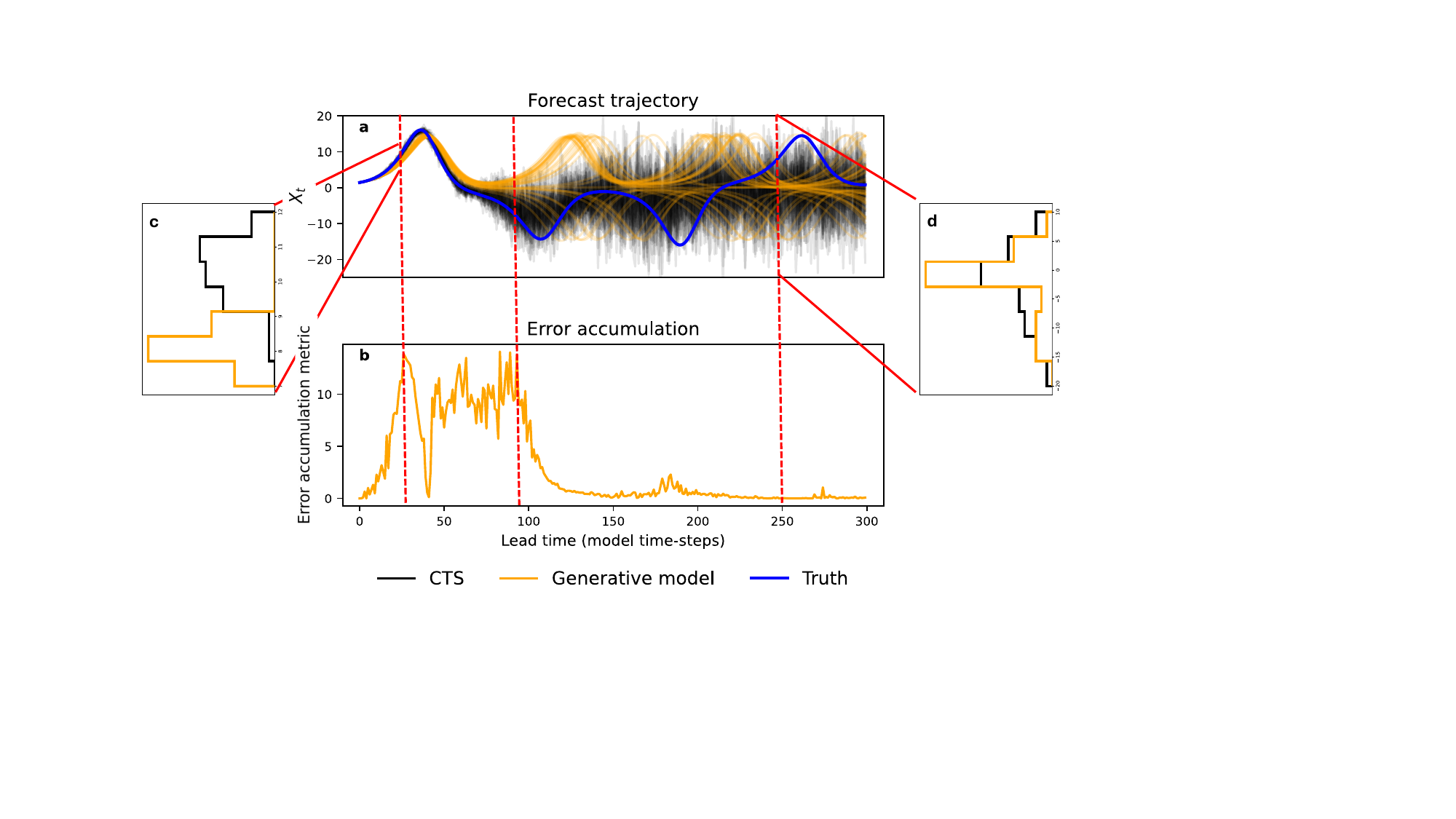}
  \caption{Illustration of error accumulation metric in equation \ref{eq:err_acc_kl_approx}. (a) 40 random simulations from a generative model and a CTS for the Lorenz 63 system, given an initial condition. (b) Error accumulation metric. It is high at time 26 as the generative model places much density outside that of the CTS, as shown in (c). It remains high at time 93 due to the generative model's inadequate coverage compared to the CTS, which centres more density around the truth (blue). The metric is low at time 250 since the generative model and the CTS distributions align well. The metric indicates the need for better generative model performance up to lead times of 100, where the CTS is more capable.} 
  \label{fig:err_acc_metric_example}
\end{figure}

In Figure \ref{fig:err_acc_metric_example},  we illustrate how the error accumulation metric changes over time for a Lorenz 63 forecast (a continuation of the ones shown in Figures \ref{fig:err_acc_non_explosion} and \ref{fig:err_acc_stic_explosion}). The metric highlights errors in the generative model that need correction. It is high initially, as iterative rollout errors cause the generative model to diverge from the CTS, indicating potential predictability gains to be made here. After time-step 200, the metric is low, suggesting that while errors are undoubtedly still created at each iterative step, their impact becomes negligible due to the loss in predictability from  chaos at these lead times. 

Training an ML model to evaluate another model's performance, as we have done here, is common in the literature. For instance, in variational autoencoders \citep{kingma2013auto,rezende2014stochastic}, an encoder is trained to approximate a posterior distribution; a better encoder brings the computed evidence lower-bound (ELBO) of the generative model closer to the log-likelihood. Similarly, in training large language models using reinforcement learning from human feedback \citep{christiano2017deep,stiennon2020learning,ziegler2019fine}, a reward model,  trained from human feedback, evaluates the language model.

KL divergence is asymmetric, and we choose the direction in equation \ref{eq:err_acc_kl} such that it is significantly larger (worse) if $p_{\mathrm{gen}}$ assigns high probability density to regions where $p_{\mathrm{cts}}$ assigns low density, rather than the reverse scenario. This case would mean the generative model, which can accumulate errors, is likely to simulate values considered very unlikely by the CTS, which does not accumulate errors. These unlikely values are penalized more strongly.  An example of this is provided in Figure \ref{fig:kl_div_explainer}. 

An alternative approach to directly compute 
equation \ref{eq:err_acc_kl} is discussed in Section \ref{appendix:fgan}. 

\subsection{Calculating the metric in practice}
\label{subsection:metric_in_practice}

Explicitly calculating the KL divergence in equation \ref{eq:err_acc_kl_approx} is typically intractable due to the required integrations. Instead, empirical procedures can be used. First, data must be sampled from $p_{\mathrm{gen}}(x_{t+c}|x_{1:c})$ since autoregressive models generally lack an explicit density function for this conditional marginal. Following this, the simplest approach, and what we use for our examples, is to approximate the densities of $x_{t+c}|x_{1:c}$ by fitting normal distributions and then applying the explicit form of KL divergence for normal distributions. This procedure performs poorly for multi-modal distributions. A more sophisticated approach to address this is to bin samples from  $x_{t+c}|x_{1:c}$ into histograms and calculate the KL divergence between them. 

\subsection{Illustrative examples}

\paragraph{Capturing explosive forms.} Returning to the Lorenz 96 example in Figure \ref{fig:err_acc_explosion}, we now introduce a basic CTS (reflecting the climatological distribution) as our reference:
\begin{equation}
    X_{k,t} = 2.5 + 6.25Z_{k,t}
    \label{eq:cts_l96}
\end{equation}

With this model, there is a 95\% chance that a sampled value lies in the range $[-10,15]$. The explosive trajectories from the generative model in Figure \ref{fig:err_acc_explosion}a are captured up by our error accumulation metric in Figure \ref{fig:err_acc_explosion}d, which  grows over the latter forecast times. This occurs because the generative model trajectories start to occupy regions of state space ($X_{3,t} < -10$ and $X_{3,t} > 15$) which the CTS allocates very little probability density too. The metric falls over the first few time-steps because our constructed CTS in equation \ref{eq:cts_l96} is not a good model for short lead times; the generative model's marginal distribution is tighter. Ideally, with a better CTS, both marginal distributions would match, resulting in a small error accumulation metric at those lead times. 

\paragraph{Capturing non-explosive forms.} In Figure \ref{fig:err_acc_non_explosion}, our metric in Figure \ref{fig:err_acc_non_explosion}d highlights increasing errors up to time-step 26, due to the difference between the generative model's inability to appropriately model the truth, and the CTS model's better tracking of it. This suggests the generative model's errors are due to model deficiencies, which we would hope to correct. 

\paragraph{Not capturing errors due to STIC.} As desired, our metric does not flag errors due to STIC, as shown in Figure \ref{fig:err_acc_stic_explosion}. Since the CTS model is not susceptible to error accumulation, the growth in its errors must be due to other factors, like STIC in this example. The generative model's errors grow similarly to the CTS, so our error accumulation metric (Figure \ref{fig:err_acc_stic_explosion}d) remains low, indicating these errors are not ones we aim to fix.

\paragraph{Not capturing errors due to unobserved variables.} Similarly, the CTS forecasts may also error growth due to a lack of predictability from unobserved variables. If the generative model's errors grow similarly, our error accumulation metric will not flag these as errors to be corrected.

%% file: chapter4/document.tex
\section{Regularization strategy}
\label{section:solution}

We propose adding a regularization penalty based on equation \ref{eq:err_acc_kl_approx} to the classic maximum likelihood objective, optimizing as follows:
\begin{equation}
\argmax_\theta \bigg( \mathbb{E}_{x_{1:n} \sim p_\mathrm{truth}} \,\mathrm{log} \,p_{\mathrm{gen},\theta}(x_{c+1},...,x_{c+n}|x_{1:c}) - \frac{\lambda}{n}\sum_{t=1}^{n} \mathbb{E}_{x_{1:c} \sim p_\mathrm{truth}} \,  \mathrm{KL}\big(p_{\mathrm{gen},\theta}(x_{t+c}|x_{1:c}) \, || \, p_{\mathrm{cts}}(x_{t+c}|x_{1:c})\big) \bigg) 
\label{eq:objective_original}
\end{equation}
where $\lambda$ is a tunable hyperparameter, and we train $p_{\mathrm{gen},\theta}$.

\paragraph{Practical implementation.} Using equation \ref{eq:objective_original} for training is challenging. First, as noted in Section \ref{subsection:metric_in_practice}, computing the KL divergence analytically is intractable. Second, even an empirical approach requires sampling from the generative model to approximate $p_{\mathrm{gen},\theta}(x_{t+c}|x_{1:c})$, which is time and memory intensive due to the need to store gradients throughout the sequence, similar to issues faced in rollout training. 

Therefore, we make the following changes to the strategy in this work. First, we modify the objective and optimize:
\begin{multline}
\argmax_\theta \bigg( \mathbb{E}_{x_{1:n} \sim p_\mathrm{truth}} \, \mathrm{log} \,p_{\mathrm{gen},\theta}(x_{c+1},...,x_{c+n}|x_{1:c})  \\  - \frac{\lambda}{n}\sum_{t=1}^{n} \mathbb{E}_{x_{1:c},x_{t+c-1} \sim p_\mathrm{truth}} \, \mathrm{KL}\big(p_{\mathrm{gen},\theta}(x_{t+c}|x_{t+c-1}) \, || \, p_{\mathrm{cts}}(x_{t+c}|x_{1:c})\big) \bigg) 
\label{eq:objective_edit}
\end{multline}
Second, following \citeauthor{sanchez2020learning}, we corrupt sections of our input data with random-walk noise during training. Third, we use normal distributions to model $p_{\mathrm{gen},\theta}(x_{t+c}|x_{t+c-1})$ and $p_{\mathrm{cts}}(x_{t+c}|x_{1:c})$. 

The penalty in equation \ref{eq:objective_edit} is not meant to be minimized to zero and does not define error accumulation directly. However, the asymmetrical nature of the KL divergence means this term penalizes $p_\mathrm{gen}$ more if it assigns density outside $p_\mathrm{cts}$, which we find empirically useful for regularization. The noise introduced to corrupt input data aims to emulate the data encountered after rolling out forecasts, bringing the penalty term in equation \ref{eq:objective_edit} closer to that in \ref{eq:objective_original}, as detailed in Section \ref{appendix:using_noise}. Using normal distributions allows analytic calculation of the KL divergence.

\paragraph{Future directions.} Instead of adding noise to emulate rollout trajectories, techniques such as the replay buffer in FengWu \citep{chen2023fengwu} could better match the effects of rollouts. Additionally, rather than relying on normal distributions to simplify KL calculations, adversarial training can be used to estimate the KL divergence term in the optimization (Section \ref{appendix:fgan_training}).


%% file: chapter5/document.tex
\section{Experiments}
\label{section:experiments}
We assess our regularization strategy on various systems, comparing it against autoregressive generative models and our trained CTS. We perform an ablation study on the Lorenz 96 to evaluate each component of our strategy. These experiments also illustrate our error accumulation metric in action. 

For our plots, we use 95\% confidence bands to show variability due to the initial conditions chosen in the test set. These bands were created using bootstrapping. Plots demonstrating the temporal consistency of autoregressive models trained using our regularization compared to CTS are shown in Section \ref{appendix:additional_results}. Further details on the systems and specific architectures are given in Sections \ref{appendix:systems} and \ref{appendix:ex_details}. 

\subsection{Lorenz 63}

\begin{figure}[h!]
  \centering
  \includegraphics[height=5.5cm]{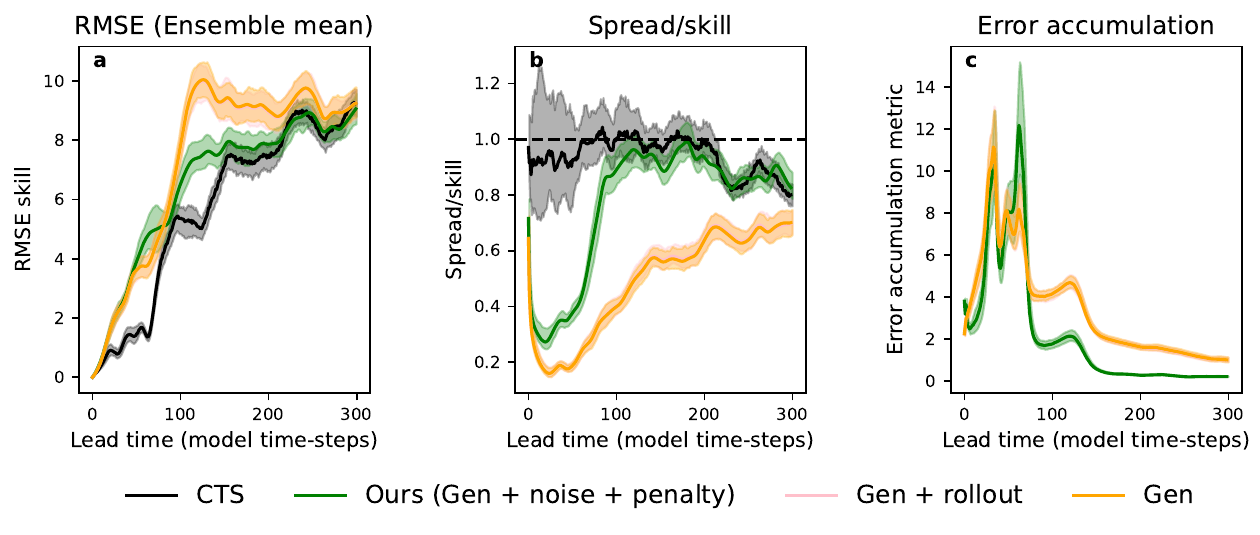}
  \caption{Evaluation of Lorenz 63 ensembles from a CTS, a generative model, a generative model with rollout training, and a generative model with our regularization strategy. (a) Ensemble-Mean RMSE skill (lower is better). (b) Spread/skill ratio (closer to 1 is better, lower suggests under-dispersion. (c) Error accumulation metric (lower is better). Our approach improves the spread/skill ratio, and achieves skill closer to the CTS at longer lead times. 95\% confidence bands are shown.}
  \label{fig:l63_main_result}
\end{figure}

We assess the predictive skill of our approach using the RMSE of the ensemble mean in Figure \ref{fig:l63_main_result}a. Our method outperforms the other generative models, including one trained with a rollout strategy, in 74\% of lead times, and in 100\% of lead times between time-steps 100 to 220. It also matches the performance of the CTS after 200 time-steps. Figure \ref{fig:l63_main_result}b measures the reliability of the forecasts. Our spread/skill score is improved (closer to one) relative to the generative models, and it is better at 100\% of lead times, though it remains under-dispersed. As expected, the CTS model has the best skill and spread/skill, given it is optimized for these metrics. The generative model with rollout training shows similar results to the one without rollout training. Figure \ref{fig:l63_main_result}c shows our error accumulation metric. The larger values for lead times less than 100 is consistent with how these models are notably worse than the CTS model for this period. Our metric therefore succeeds in diagnosing errors we aim to fix. At longer lead times, the metric drops as the models approach the CTS performance. Although RMSE increases, this is  expected as we reach the predictability horizon. The drop in our metric indicates these may not be fixable errors. Our approach drops to near zero after lead time 200, reflecting how its RMSE and spread/skill match the CTS model after this time. Our approach remains stable for long-run simulations  beyond the 150 time-steps upon which it is trained.

\subsection{Ablation study for the Lorenz 96}

\begin{figure}[h!]
  \centering
  \includegraphics[height=5.5cm]{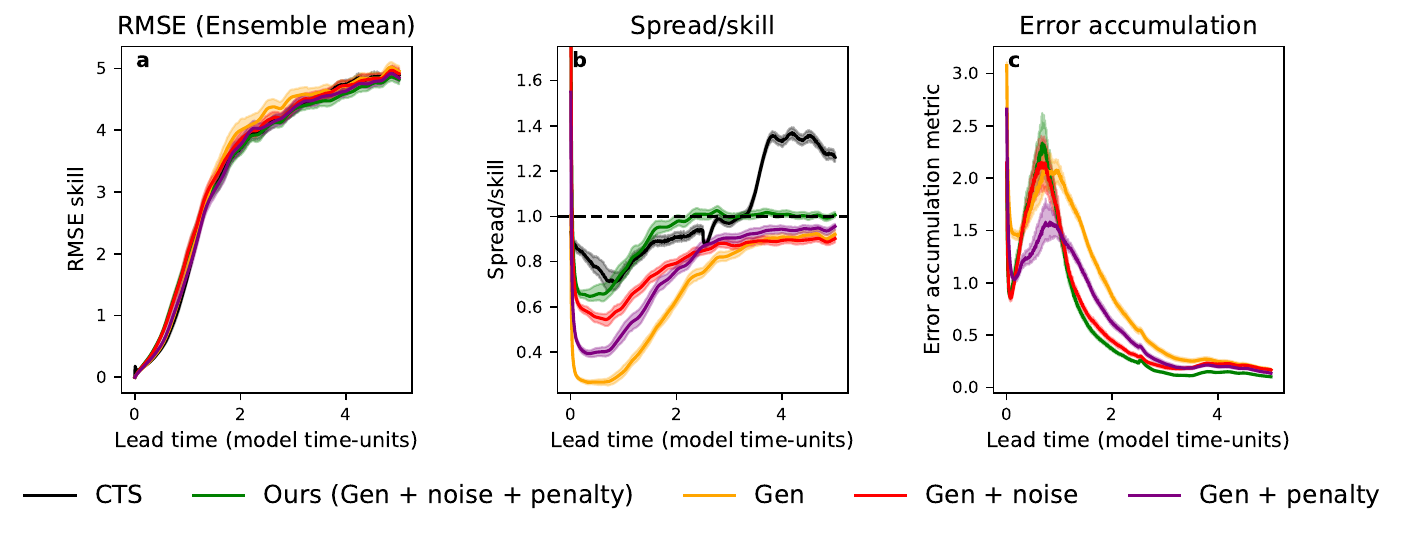}
  \caption{Ablation study (Lorenz 96) examining the contributions of noise and the KL penalty to our regularization strategy. (a) Ensemble-Mean RMSE skill (lower is better). (b) Spread/skill ratio (closer to 1 is better, lower suggests under-dispersion. (c) Error accumulation metric (lower is better). Both components are important for improving performance at longer lead times. 95\% confidence bands are shown.}
  \label{fig:l96_main_result}
\end{figure}

In Figure \ref{fig:l96_main_result}, we observe the importance of both noise and the KL penalty in our regularization strategy. The RMSE for all approaches in Figure \ref{fig:l96_main_result}a are  similar. However, in Figure \ref{fig:l96_main_result}b, the spread/skill improves as the base generative model is trained with the KL penalty alone, with noise alone, and with both combined. Our approach outperforms using the individual components at 100\% of lead times for spread/skill, and is the only one achieving a spread/skill of one. The error accumulation metric in Figure \ref{fig:l96_main_result}c shows a similar trend, though it is the generative model using only the penalty which performs best at lead times around 1 model time-unit. Our approach (green) performs better at lead times greater than 2 model time-units. Despite the CTS having deficiencies with poor spread/skill at later lead times, using it in the KL penalty remains beneficial.

Further qualitative results are provided in Section \ref{appendix:l96_figure}.

\subsection{Weather Prediction}

\begin{figure}[h!]
  \centering
  \includegraphics[height=5.5cm]{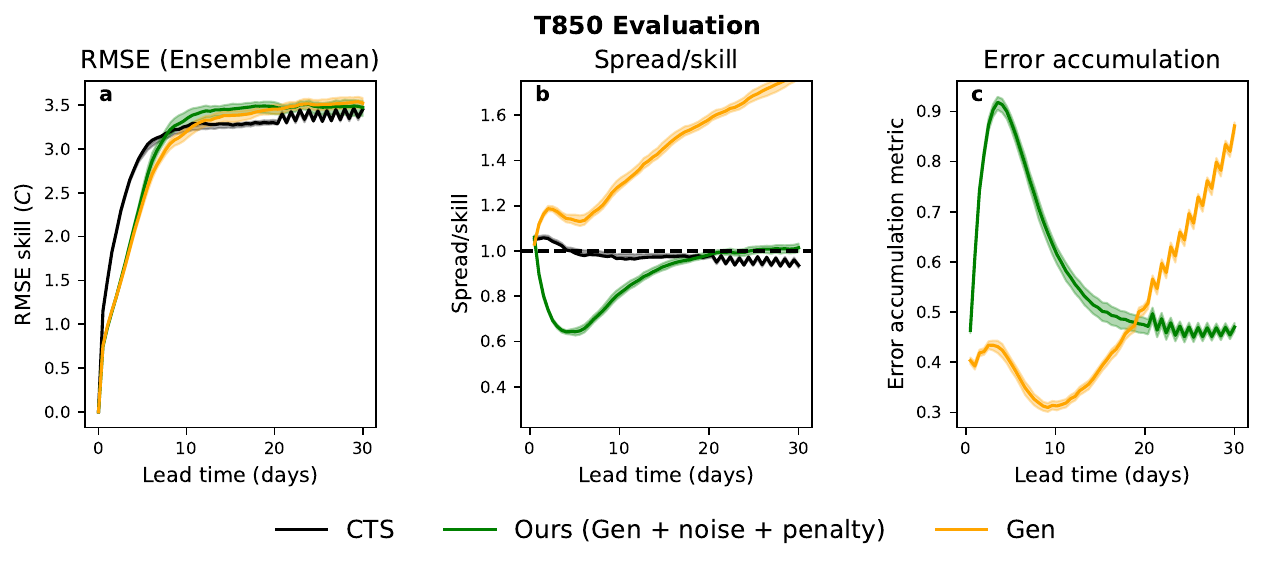}
  \caption{Evaluation of the ensembles produced by a CTS, a generative model and a generative model with our regularization strategy for T850. (a) Ensemble-Mean RMSE skill (lower is better). (b) Spread/skill ratio (closer to 1 is better, lower suggests under-dispersion. (c) Error accumulation metric (lower is better), detecting issues in the generative model due to continuing increase in spread. 95\% confidence bands are shown.}
  \label{fig:nwp_t850}
\end{figure}

In Figure \ref{fig:nwp_t850}, we evaluate T850 (temperature at pressure level 850), a key metric reported in MLWP. Additional results for Z500 and MSLP are in Section \ref{appendix:additional_results}, as well as forecast visualizations. Notably, Figure \ref{fig:nwp_t850}b shows that the spread of the generative model continues to increase even though the RMSE (Figure \ref{fig:nwp_t850}a) remains stable, suggesting potential explosions in particular samples. This is captured in Figure \ref{fig:nwp_t850}c, as intended. 

The jagged pattern in the CTS is due to our modelling approach: for $t>20$, CTS samples $x_t$ are taken from $p_\mathrm{cts}(x_{c+20}|x_c)$. While this is a fair approximation given that weekly climatology is reached after 20 days, it does not account for the diurnal cycle, causing the jagged patterns.

The CTS requires significant improvement. In Figure \ref{fig:nwp_t850}, it performs worse than both the other models for lead times up to 10 days. Moreover, in Section \ref{section:diagnostic}, we diagnose why the error accumulation metric fails to capture explosion in Figures \ref{fig:nwp_z500}--\ref{fig:nwp_mslp}, and show this can be traced back to issues in the CTS.

\section{Discussion}
\label{section:limitations}


\paragraph{Limitations} 
Our definition in equation \ref{eq:err_acc_kl} is challenging to compute, necessitating approximations such as equation \ref{eq:err_acc_kl_approx}. As noted in  Section \ref{section:def}, training a model to evaluate another one is common practice. Whilst a better CTS yields a better approximation of equation \ref{eq:err_acc_kl}, our experiments show that even a CTS which performs worse than an autoregressive model for certain lead times can still be useful. In our regularization strategy, simplifications such as using noise to emulate rollouts and normal distributions were necessary to make equation \ref{eq:objective_original} tractable. Again, the approach depends on the quality of the CTS. Nevertheless, if the approximations result in an unhelpful penalty term, the hyperparameter $\lambda$ can be adjusted during  tuning to account for this. Sections \ref{appendix:fgan} and \ref{appendix:fgan_training} suggest more sophisticated approaches to reduce the need for approximations. 


\paragraph{Conclusions} We propose a definition of error accumulation and associated metric which can diagnose fixable errors. We suggest that an effort to create a good CTS at higher spatio-temporal resolutions will support model development across research groups. Finally, we hope that our error accumulation definition will inspire further solutions.


%% file: appendix.tex
\section{Broader impacts}
\label{appendix:broader_impacts}

This work contributes to the wider mission of improving NWP. Using ML enables forecasts to be run orders of magnitude quicker, allowing more ensembles to be run, and making it easier for more downstream users to use weather forecast output. Our work aims to contribute to the task of decreasing ML model deficiencies. More accurate models are more useful for decision-making for a variety of tasks, such as renewable energy grid optimization.

The main negative impact of the community's work to improve MLWP is that training operational-scale models requires significant time and compute resources, which will result in notable greenhouse gas emissions. For example, GraphCast \citep{lam2022graphcast} took four weeks to train on 32 TPU devices. It is also important to note that most MLWP are currently fully data-driven. These are not causal models (unlike standard NWP which are based on physical equations), so their results must be interpreted with care, especially when extrapolating. 

\section{Existing metrics}

In the following,

\begin{itemize}
    \item $x^n_{i,m}$ is the value of the $n$th of $N$ ensemble members on the $m$th of $M$ forecasts for a given variable, level and lead time, at latitude and longitude indexed by $i \in G$.
    \item $\Bar{x}_{i,m} = \frac{1}{N} \sum_n x^n_{i,m}$ denotes the ensemble mean.
    \item $x^O_{i,m}$ denotes the corresponding target ($O$ stands for `observed'). 
    \item $a_i$ denotes the area of the latitude-longitude grid cell. This varies by latitude and is normalized to unit mean over the grid.
    
\end{itemize}

\subsection{RMSE}
\label{appendix:rmse}

The ensemble mean RMSE is defined as:

\begin{equation}
    \textrm{RMSE} := \sqrt{\frac{1}{M}\sum_m  \frac{1}{|G|}\sum_i a_i (\Bar{x}_{i,m} - x^{O}_{i,m})^2}
    \label{eq:rmse}
\end{equation}

\subsection{Spread/skill}
\label{appendix:spread_skill}
Spread is defined as the root mean estimate of ensemble variance:

\begin{equation}
    \textrm{Spread} := \sqrt{\frac{1}{M}\sum_m \frac{1}{|G|} \sum_i a_i \frac{1}{N} \sum_n (x^n_{i,m} - \bar{x}_{i,m})^2 }
\end{equation}

Skill is defined as the ensemble mean RMSE from equation \ref{eq:rmse}. Spread/skill ratio is defined as:

\begin{equation}
    \textrm{Spread/skill} := \frac{\textrm{Spread}}{\textrm{Skill}}
    \label{eq:spread_skill}
\end{equation}
and, as noted by \citeauthor{leutbecher2008ensemble}, this should be 1 for a perfectly reliable forecast (defined as one where $x^n_{i,m}$ and $x^O_{i,m}$ are independent samples from the same distribution).

\subsection{CRPS}
\label{appendix:crps}

The CRPS is defined as:

\begin{equation}
\textrm{CRPS} := \frac{1}{M} \sum_m \frac{1}{|G|} \sum_i a_i \int_\mathbb{R} \big(F_{X_{i,m}}(x) - H(x \geq x^O_{i,m})\big)^2 dx
\end{equation}

where $F_{X_{i,m}}$ is the cumulative distribution function for the modelled distribution $X_{i,m}$ and $H$ is the Heaviside step function. Intuitively, the CRPS can be seen as the CDF version of the likelihood function. The likelihood function measures the likelihood (the probability density) assigned to observed data. The CDF measures how much probability mass is concentrated around the observed data. 

\section{Error accumulation metric}

\subsection{Links to other metrics}

With the RMSE, although a lower value is desirable, there is a natural upper bound  imposed by properties of the system (chaos, unobserved variables), so there is no clear reference point. The spread/skill ratio, on the other hand, has a clear reference point of one, indicating perfect reliability. However, a model with signficant errors can still achieve a perfect spread/skill score if its spread is sufficiently large. While spread/skill captures forecast reliability, it does not distinguish between forecasts with identical spread/skill but different skill levels. Therefore, it becomes necessary to combine the results of multiple metrics, but as shown in Section \ref{section:problem_formulation}, this approach complicates drawing conclusions related to fixable error accumulation.

The CRPS can be understood to measure qualities relating to both the RMSE and spread of a forecast. However, these are measured relating to the truth data, so yet again, there is no clear reference point: although a CRPS of zero is ideal, this may not always be achievable due to inherent system properties. 

\subsection{Asymmetry of KL divergence}

\begin{figure}[h!]
  \centering
  \hspace*{-1.1cm}
  \includegraphics[height=4cm]{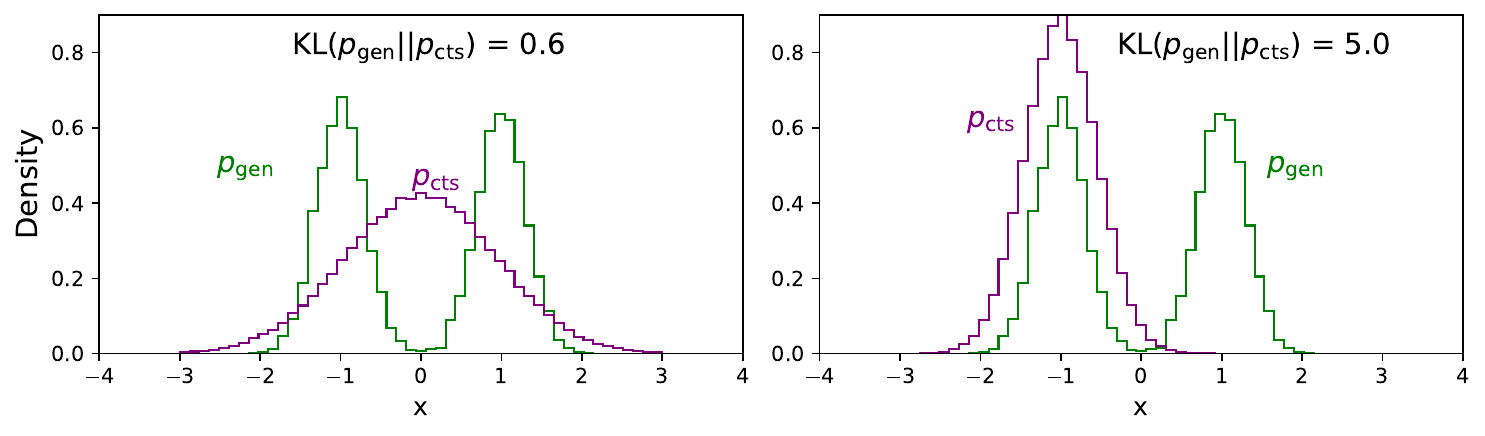}
  \caption{Illustration of the asymmetry of the KL divergence. The KL divergence is far greater when the generative model assigns density to regions which are assigned low density by the CTS.} 
  \label{fig:kl_div_explainer}
\end{figure}

Figure \ref{fig:kl_div_explainer} illustrates the asymmetry of the KL divergence, motivating the reason for the particular choice of direction in our metric. 

\subsection{Alternative approach to directly compute equation \ref{eq:err_acc_kl}}
\label{appendix:fgan}

An alternative approach is to train a model to approximate the KL divergence in equation \ref{eq:err_acc_kl}, as done for f-GANs \citep{nowozin2016f}. In this work, we opt to use the CTS formulation instead, as it requires less computational cost during training. The f-GAN approach requires sampling from $p_\mathrm{gen}$ during training, which, while feasible (as in common rollout training strategies), is more complex than training a CTS model. Moreover, using the CTS is more interpretable. We can evaluate it using standard metrics to assess how good it is --- we would wish it to at least be better than a generative model at capturing conditional marginals, given that is what it is trained for. In contrast, with the f-GAN approach, it is not clear how good the resulting approximation for the KL divergence is. Only for the optimal critic is the KL divergence perfectly captured. 

\section{Regularization training approach}

\subsection{Links to the rollout training approach}

The rollout training approach in the literature can be understood as optimizing:
\begin{equation}
\mathbb{E}_{x_{1:n} \sim p_\mathrm{truth}}\bigg[\frac{1}{n} \sum_{t=1}^n \log p_{\mathrm{gen},\theta}(x_{t+c}|x_{1:c})\bigg]
\end{equation}
with respect to $\theta$. This can be linked to a KL divergence as follows:

\begin{align}
\mathbb{E}_{x_{1:n} \sim p_\mathrm{truth}}\bigg[\frac{1}{n} \sum_{t=1}^n \log p_{\mathrm{gen},\theta}(x_{t+c}|x_{1:c})\bigg] =- \frac{1}{n} \sum_{t=1}^n  \mathbb{E}_{x_{1:c} \sim p_\mathrm{truth}}  \operatorname{KL}\big(p_{\mathrm{truth}}(x_{t+c}|x_{1:c}) \, || \,p_{\mathrm{gen},\theta}(x_{t+c}|x_{1:c})\big)  + k
\end{align}
where $k$ is a constant. The rollout training approach therefore involves doing:

\begin{equation}
\argmax_\theta \bigg( - \frac{1}{n} \sum_{t=1}^n \mathbb{E}_{x_{1:c} \sim p_\mathrm{truth}}  \operatorname{KL}\big(p_{\mathrm{truth}}(x_{t+c}|x_{1:c}) \, || \,p_{\mathrm{gen},\theta}(x_{t+c}|x_{1:c})\big)\bigg) 
\label{eq:rollout}
\end{equation}

Our approach in equation \ref{eq:objective_original} differs in how we i) include the standard maximum likelihood term, which encourages correlations to be captured, ii) use the KL term as a penalty as opposed to the ultimate objective, and iii) use a direction of the KL divergence that penalizes the generative model more if it puts density outside the CTS. 


\subsection{Using noise to emulate rollouts}
\label{appendix:using_noise}

When using noise, we create our noisy data using $\tilde{x}_{t+c-1} = x_{t+c-1} + Z_t$, where the noise process $Z_t$ is specified in Appendix \ref{appendix:ex_details}. We hope that $p_{\mathrm{gen},\theta}(x_{t+c}|\tilde{x}_{t+c-1})$ is a sufficiently good enough match to $p_{\mathrm{gen},\theta}(x_{t+c}|x_{1:c})$ to enable useful training. 

Our motivation is as follows. Explicitly calculating $p_{\mathrm{gen},\theta}(x_{t+c}|x_{1:c})$ would require doing the intractable integral:

\begin{equation}
    p_{\mathrm{gen},\theta}(x_{t+c}|x_{1:c}) = \int_{x_{c:t+c-1}}  p_{\mathrm{gen},\theta}(x_{t+c},x_{c:t+c-1}|x_{1:c}) \, dx_{c:t+c-1} 
\end{equation}

This can be simplified using the properties of the graphical model we are using to:

\begin{equation}
    p_{\mathrm{gen},\theta}(x_{t+c}|x_{1:c}) = \int_{x_{t+c-1}}  p_{\mathrm{gen},\theta}(x_{t+c}|x_{t+c-1}) \, p_{\mathrm{gen},\theta}(x_{t+c-1}|x_{1:c}) \, dx_{t+c-1} 
\end{equation}
and this can be approximated with Monte-Carlo sampling with:
\begin{equation}
    p_{\mathrm{gen},\theta}(x_{t+c}|x_{1:c}) \approx \frac{1}{N} \sum_{i=1}^{N}p_{\mathrm{gen},\theta}(x_{t+c}|x^i_{t+c-1}), \, \,\,\, x^i_{t+c-1} \sim p_{\mathrm{gen},\theta}(x_{t+c-1}|x_{1:c})
\end{equation}

Now, sampling from $p_{\mathrm{gen},\theta}(x_{t+c-1}|x_{1:c})$ would require running the model forward in time, which is computationally expensive. Instead, as is done by \citeauthor{sanchez2020learning}, we use noise to mimic the error-accumulating results of the sampling process, setting $x^i_{t+c-1} = \tilde{x}^i_{t+c-1}$. 

We suspect that using a noise scheduler to decrease the amount of noise which is added as training time increases would improve results. This would better mimic the fact that better trained models would undergo less error accumulation. However, this was not examined further in this work.

\subsection{Training with f-GANs}
\label{appendix:fgan_training}

The optimization with respect to $\theta$ of the penalty term in equation \ref{eq:objective_original} can be carried out with f-GANs \citet{nowozin2016f}. In the approach, one model is trained to estimate the KL divergence, and then the generative model is updated to minimize the KL divergence. This can then be combined with a maximum likelihood objective to directly optimize equation \ref{eq:objective_original}. We use the CTS approach instead since it is simpler to train, and avoids the challenges from doing min-max optimization in GANs. Nevertheless, we note that the f-GAN approach would be an interesting direction to pursue.

\section{Atmospheric systems}
\label{appendix:systems}

In our released code, we provide scripts to create (or download) and process the required data for each of these systems. 

\subsection{Lorenz 63}

The Lorenz 63 system \citep{lorenz1963deterministic} is a system of ordinary differential equations, famous for its use in studying chaos. The equations are

\begin{align}
    \frac{dx}{dt} &= \sigma (y-x) \\ 
    \frac{dy}{dt} &= x(\rho - z) - y \\
    \frac{dz}{dt} &= xy - \beta z
\end{align}
and we set the parameters $\sigma = 10 $, $\rho = 28 $ and $\beta = 2.667 $. The equations are solved using a fourth order Runge-Kutta scheme with a time-step of $\delta t = 0.01 $ to create $1,500,000$ time-steps of data. The first $200$ are discarded. The first $900,000$ are used for training, the next $100,000$ for validation, and the remaining $500,000$ for testing.

We set ourselves the task of modelling $X_t$ and $Y_t$ conditioned on a single initial condition. The omission of $Z_t$ means that we are dealing with unobserved variables.

\subsection{Lorenz 96}

We use the two-tier Lorenz 96 model \citep{lorenz1996predictability}, a toy model for atmospheric circulation that is extensively used for atmospheric parameterization studies \citep[e.g.][]{hannah_bespoke,crommelin2008subgrid,gan_hannah,kwasniok2012data,rasp2020}. We use the configuration described in \citet{gan_hannah}. It comprises two types of variables: large, low-frequency variables, $X_k$, arranged in a circle, interacting with small, high-frequency variables, $Y_{j}$. These are dimensionless quantities, evolving as follows:
\begin{equation*}
    \frac{dX_k}{dt} =\underbrace{ -X_{k-1}(X_{k-2} - X_{k+1})}_{\textrm{advection}} \underbrace{- X_{k}}_{\textrm{diffusion}} \ \underbrace{+ F}_{\textrm{forcing}} - \underbrace{\frac{hc}{b} \sum_{j=J(k-1)+1}^{kJ} Y_j}_{\textrm{coupling}}  \mkern70mu k = 1, ..., K 
\end{equation*}
\begin{equation*}
    \frac{dY_j}{dt} = \underbrace{-cbY_{j+1}(Y_{j+2} - Y_{j-1})}_{\textrm{advection}} \: \underbrace{- \, cY_{j}}_{\textrm{diffusion}} \: \underbrace{ - \frac{hc}{b} X_{int[(j-1)/J]+1}}_{\textrm{coupling}}   \mkern105mu j = 1, ..., JK 
\end{equation*}
where the variables have cyclic boundary conditions: $X_{k+K}=X_{k}$ and $Y_{j+JK} = Y_j$. The $X$ variables represent the atmospheric conditions along a latitude circle around a hypothetical planet. In our experiments, the number of $X$ variables is $K=8$, and the number of $Y$ variables per $X$ is $J=32$. The value of the constants are set to $h=1,\:b=10$, $\:c=10$ and $F\:=20$. These indicate that the fast variable evolves ten times faster than the slow variable and has one-tenth the amplitude. The chosen parameters follow those used in \citet{hannah_bespoke}, and are such that one model time unit (MTU) is equivalent to five days in the Earth's atmosphere. This equivalence follows from the error doubling time for the model being comparable to that in the atmosphere over these time-scales \citep{lorenz1996predictability}.

The truth data was created by solving the equations using a fourth-order Runge-Kutta timestepping scheme and a fixed time step $\Delta t = 0.001$ MTU, with the output saved at every 0.005 MTU. A dataset of length 45000 MTU (9,000,000 time-steps). The first 1,000,000 time-steps were used for training, the next 100,000 for validation, and the remaining kept for testing.

In this work we model the $X$ variables, treating the $Y$ as unobserved. The coupled set of Lorenz 96 equations are treated as the `truth'. This set-up matches the common one in parameterization, with the analogy being that we can consider the $X_k$ to be coarse processes resolved in both low-resolution and high-resolution simulators, whilst the $Y_{j}$ can be considered as those that can only be resolved in high-resolution, computationally expensive simulators. 

\subsection{Weather prediction}
\label{appendix:era5_info}

As in previous studies here \citep[e.g.][]{bi2023accurate,lam2022graphcast,nguyen2023scaling,price2023gencast,rasp2021data}, we use regridded ERA5 data. ERA5 (Fifth Generation of the European Centre for Medium Range Weather Forecasts Reanalysis) \citep{hersbach2020era5} is a state-of-the-art global atmospheric reanalysis dataset. We download data at $5.625^\circ$ (32x64 grids) from WeatherBench \citep{rasp2020weatherbench} at: \url{https://github
.com/pangeo-data/WeatherBench} (MIT License). The entire dataset at this resolution is 191 GB. 

We use data from 1979 to 2017 for training, data from the year of 2018 for validation, and data from 2019 for testing. 

We choose to model eastward wind (U), northward wind (V), specific humidity (Q), geopotential (Z) and temperature (T) at 200, 500, 700 and 850 hPa pressure levels, and four surface-level variables: 2-meter temperature (T2M), 10-meter U and V components of wind (U10 and V10), and mean sea-level pressure (MSLP). Together, this results in 24 variables to track at each spatial location and time-point. We model the data at a 12-hourly resolution at 00z/12z (where 00z means 00:00 UTC) for the reasons discussed by \citeauthor{price2023gencast}. Briefly, the assimilation procedure in ERA5 results in different distributions of data based on when the assimilation took place. Choosing a 12-hour time step avoids training on this bimodal distribution. 

\section{Experimental details}
\label{appendix:ex_details}

We describe the key points below. Our code provides further implementation detail. Models were trained on a 32GB NVIDIA V100S GPU.

\subsection{Lorenz 63}

\subsubsection{Model architectures}

Our base generative model consists of a stack of 3 feedforward layers with ELU activation and dimension of 32. The CTS consists of 5 feedforward layers with ELU activation and dimension of 32. We parameterize the CTS to directly model $X_t|X_0$ for $t=1,...,200$. For $t>200$, we use $X_{200}|X_0$ for simplicity. In both cases, the residual between the current time-step and next one is modelled, as well as the sigma. 

\subsubsection{Training}

All data is standardized before training. The CTS was trained for 10 epochs using Adam \citep{kingma2014adam}, and the other models for 20 epochs. The CTS was trained on $t=1,...,200$. For the CTS model, the learning rate was 3e-4 for the base generative model and 1e-4 for the CTS. For our regularization approach, we set $\lambda=5$, and only introduce noise for the latter 10 epochs for 30\% of each batch. The learning rate is 3e-4 for the first 25 epochs, and then for the remaining 25 epochs we reduce the learning rate for the last layer which is used to output sigma to 1e-7. For our noise scheme, we used $Z_t \sim N(0,0.003t)$.

\paragraph{Compute resources} Each model trained in 1-3 hours. Approximately triple the runs shown in the paper were done, including preliminary experiments, and failed experiments.

\subsection{Testing}

For our experiments, we select 400 random initial conditions from the test data. 100 datasets were simulated for bootstrapping.

\subsection{Lorenz 96}

\subsubsection{Model architectures}

Our base generative model consists of a stack of 8 feedforward layers with ELU activation and dimension of 64. The CTS consists of 8 feedforward layers with ELU activation and dimension of 256. We parameterize the CTS to directly model $X_t|X_0$ for $t=1,...,500$. For $t>500$, we use $X_t|X_{500}$, where $X_{500}$ is sampled from the CTS. In both cases, the residual between the current time-step and next one is modelled, as well as the sigma. 

\subsubsection{Training}

All data is standardized before training. All models were trained for 50 epochs using Adam \citep{kingma2014adam}. The CTS was trained on $t=1,...,500$. For the CTS, the learning rate was 3e-5, and for the base generative model it was 1e-4. For our regularization approach, we set $\lambda=5$, and only introduce noise for the latter 25 epochs for 20\% of each batch. The learning rate is 1e-4 for the first 25 epochs, and then for the remaining 25 epochs we reduce the learning rate for the last layer which is used to output sigma to 1e-7. For our noise scheme, we experimented with different noise standard-deviations for the random walk. Empirically, we found that a greater noise than provided by the random walk for earlier time-steps was beneficial, so chose the noise process $Z_t \sim N(0,\textrm{14e-6} \, t + 0.0042)$.

\paragraph{Compute resources} Each model trained in 2-5 hours. Approximately triple the runs shown in the paper were done, including preliminary experiments, and failed experiments.

\subsubsection{Testing}

For our experiments, we select 500 random initial conditions from the test data. 100 datasets were simulated for bootstrapping.

\subsection{ERA5}

\subsubsection{Model architectures}


The backbone for all models is a Vision Transformer (ViT), similar to the one used by \cite{assran2023self}. In the CTS, the time embedding is fused into the model using AdaLN \citep{Peebles2022DiT,perez2018film}. Additionally, the CTS has a learnable sigma parameter for each target variable gor each time step. In the autoregressive models (generative and ours) we learn a sigma parameter for each target variable at the next time step. We parameterize the CTS to directly model $X_t|X_0$ for $t=1,...,20$ days. For $t>20$, we choose to sample from $X_{20}|X_{0}$, given weekly climatology is reached. For our approach, we choose to model the residual between the CTS's mean prediction of the target time step and the true value. The architecture hyperparameters are kept similar for both autoregressive models, resulting in a similar number of trainable parameters (Table \ref{tab: hp}).


\begin{table}[ht]
  \centering
  \caption{Hyperparameters for the different models used in experiments on ERA5 data}
  \begin{tabular}{lccc}
    \toprule
    \textbf{Hyperparameter} & \textbf{CTS} & \textbf{Gen} & \textbf{Ours} \\
    \midrule
    \textbf{Architecture} & & & \\
    Drop Rate (Attention) & 0.2 & 0.1 & 0.1 \\
    Drop Rate (MLP) & 0.2 & 0.1 & 0.1 \\
    Drop Rate (Path) & 0.2 & 0.1 & 0.1 \\
    Depth & 4 & 8 & 8 \\
    Embed Dimension & 128 & 512 & 512 \\
    Number of Heads & 4 & 8 & 8 \\
    Patch Size & 2 & 4 & 4 \\
    \midrule
    \textbf{Training} & & & \\
    Batch Size & 64 & 16 & 16 \\
    $\lambda$ & N/A & N/A & 0.1 \\
    Learning Rate & 0.0005 & 0.0001 & 0.0001 \\
    Max Epoch & 75 & 40 & 40 \\
    Seed & 49 & 42 & 49 \\
    \bottomrule
  \end{tabular}
  \label{tab: hp}
\end{table}

\subsubsection{Training}



All data is standardized before training. The CTS was trained on $t=1,...,20$ days. The CTS model was trained with early stopping based on validation loss. The top 3 models were saved, and the model at the earliest epoch was chosen (to prevent overfitting on the training data). For the autoregressive models, the trained model at the end of last epoch was used. We trained using AdamW  \citep{loshchilov2018decoupled} with its default settings. Further training details are given in Table \ref{tab: hp}. For our noise scheme, we used $Z_t \sim N(0,0.00001t)$. We used a number of regularization techniques in training. First, is attention dropout, which randomly sets some attention values to zero. Second, is standard dropout in the feed-forward network within each transformer block. Third, is stochastic depth (path dropout), which  drops entire transformer blocks while training.

\paragraph{Compute resources} Approximately five times the runs shown in the paper were done, including preliminary experiments, and failed experiments. The Gen and CTS models took 1-3 hours to train, while ours took roughly 2-4 hours to train. The CTS model is 14 MB with 1.3M parameters. The Gen and Our model are $\sim$ 300 MB with $\sim$ 27 M parameters.

\subsubsection{Testing}

For our experiments, we select 50 random initial conditions from the test data. 50 datasets were simulated for bootstrapping.

\section{Code}
\label{section:code}



Our code will be released following the peer-review process.

\section{Additional results}
\label{appendix:additional_results}

\paragraph{Temporal consistency} We wish to verify that the performance improvements shown by our models in Section \ref{section:experiments} are not merely due to the model collapsing to the CTS. The CTS is not temporally consistent, and this is not captured by RMSE, spread/skill, or our error accumulation metric. We can assess this qualitatively by plotting sampled trajectories.

For the Lorenz 96, in Figure \ref{fig:l96_trajectories}, our approach produces temporally consistent samples, unlike the CTS. Similarly, for the NWP task, in Figures \ref{fig:t850}--\ref{fig:10u}, we see the CTS model samples are far noisier than those from our model. This is because our approach still captures temporal correlations, unlike the CTS. This supports that our regularization approach does not simply collapse the generative model to the CTS. Rather, we are able to combine the benefits of autoregressive models (temporal consistency, and simulations of arbitrary lengths) with those of the CTS. 

\paragraph{CTS deficiencies} Although our error accumulation metric correctly detects  explosive behaviour in Figure \ref{fig:nwp_t850}, it fails to do so in Figures \ref{fig:nwp_z500}--\ref{fig:nwp_mslp}. In Section \ref{section:diagnostic}, we examine how these issues can be traced to deficiencies in the CTS.  

\subsection{Lorenz 96}
\label{appendix:l96_figure}

\begin{figure}[h!]
  \centering
  \hspace*{-1.1cm}
  \includegraphics[height=18cm]{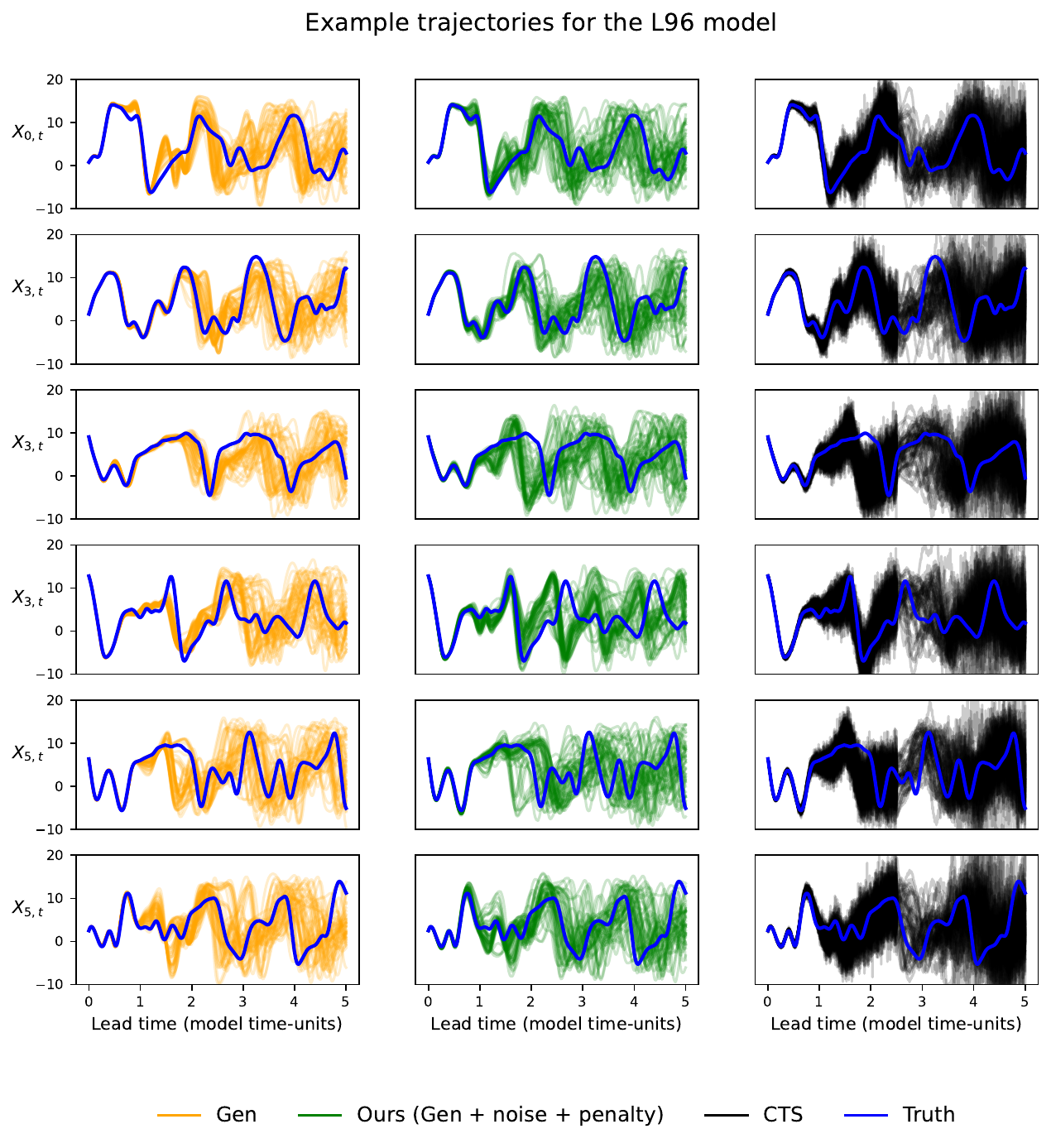}
  \caption{Example trajectories for the Lorenz 96. 50 samples were created for each trained model.} 
  \label{fig:l96_trajectories}
\end{figure}

In Figure \ref{fig:l96_trajectories}, we plot example trajectories for different spatial components. Both our approach and the base generative model (orange) give temporally consistent trajectories, unlike the CTS, as is to be expected. We note that regions where our approach does poorly often do coincide with regions of poor CTS performance. This suggests further improvement is possible with further CTS model development.

\subsection{Numerical Weather Prediction}

\begin{figure}
  \centering
  \hspace*{-1.5cm}
  \includegraphics[height=17cm]{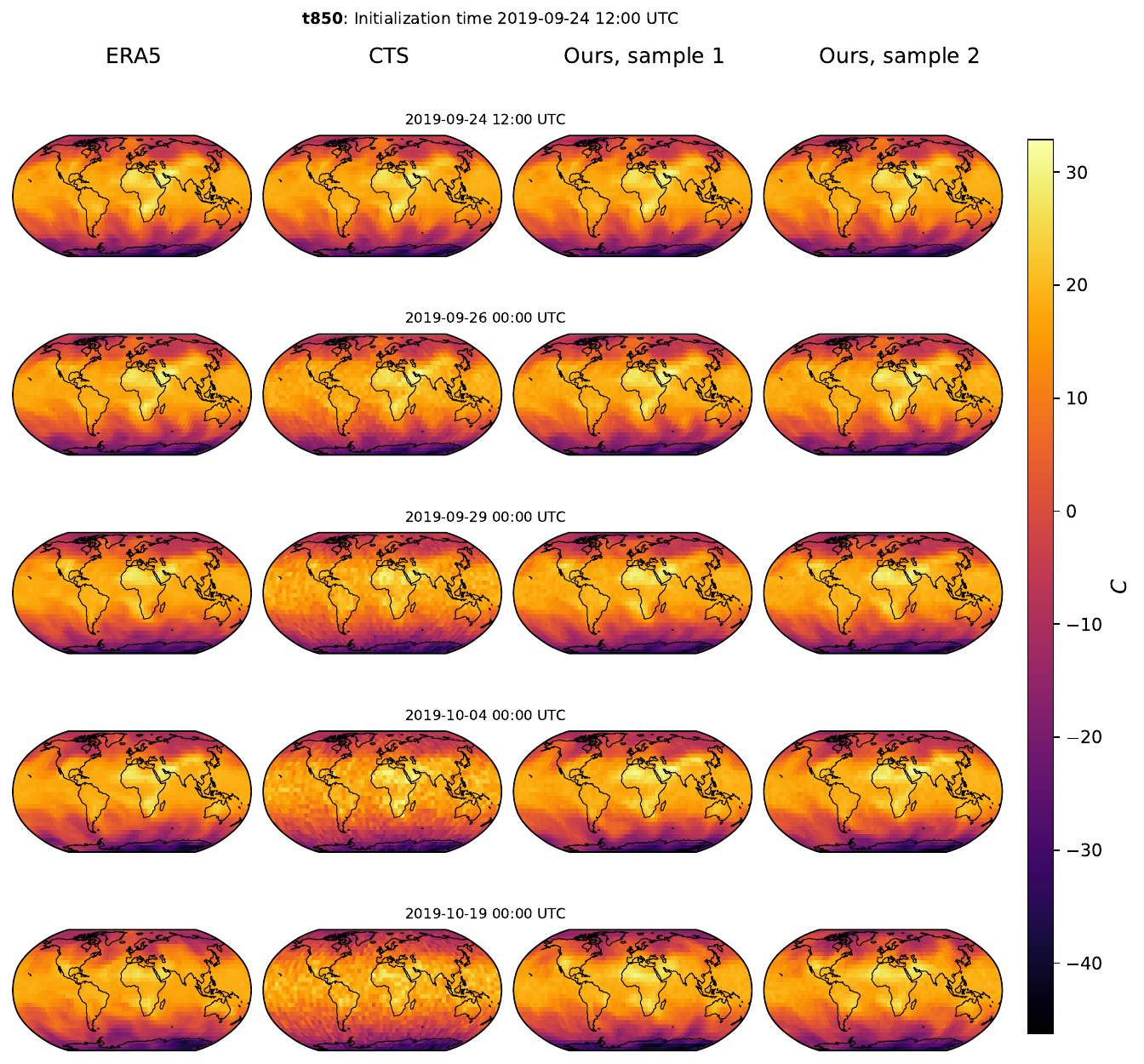}
  \caption{Forecast visualization: T850. Forecast initialized at 2019-09-24 12:00 UTC, with plots corresponding to 2, 5, 10 and 25 day lead times. Two sampled trajectories from our model are shown.} 
  \label{fig:t850}
\end{figure}

\begin{figure}
  \centering
  \hspace*{-1.5cm}
  \includegraphics[height=17cm]{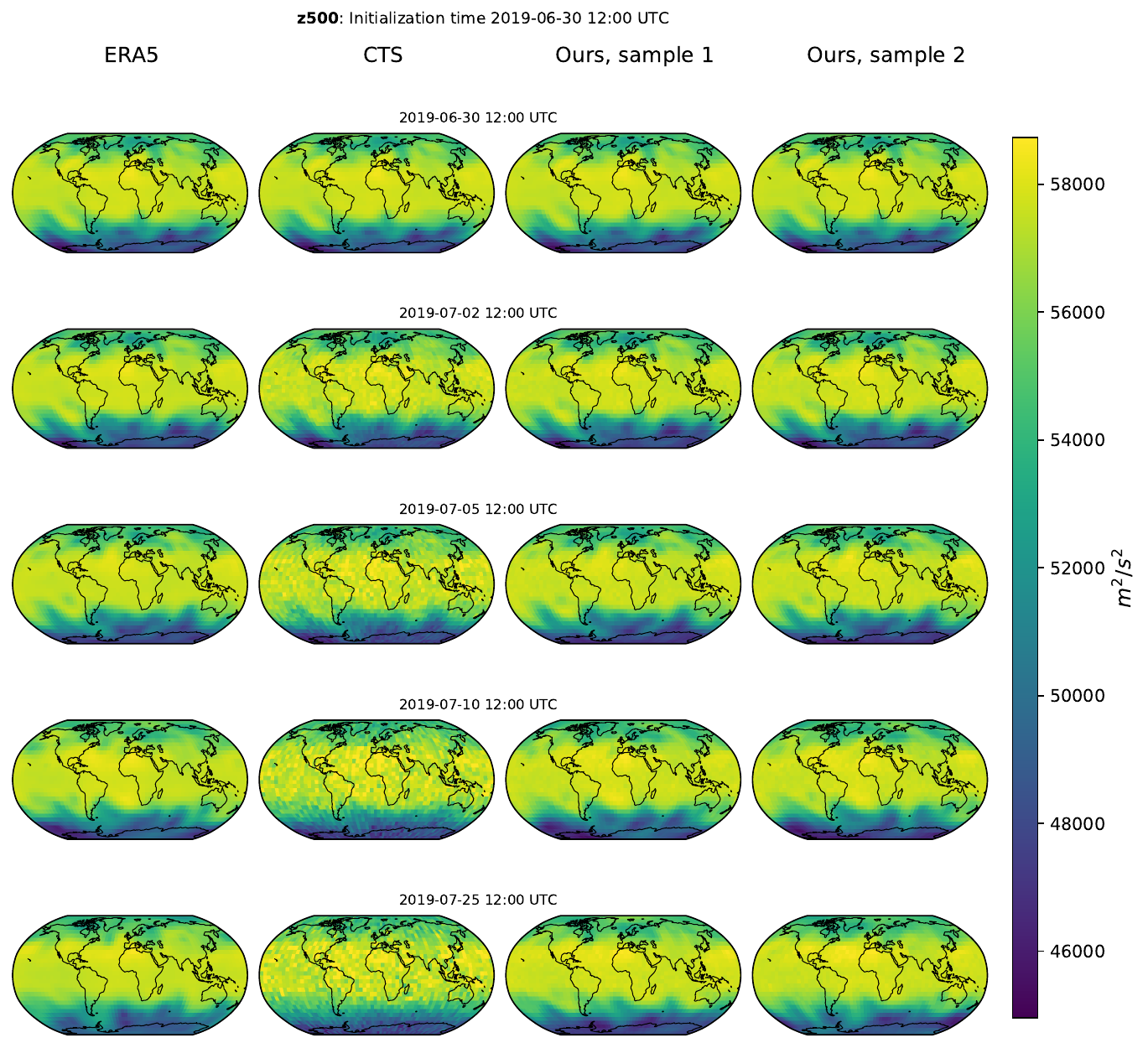}
  \caption{Forecast visualization: z500. Forecast initialized at 2019-06-30 12:00 UTC, with plots corresponding to 2, 5, 10 and 25 day lead times. Two sampled trajectories from our model are shown.} 
  \label{fig:z500}
\end{figure}

\begin{figure}
  \centering
  \hspace*{-1.5cm}
  \includegraphics[height=17cm]{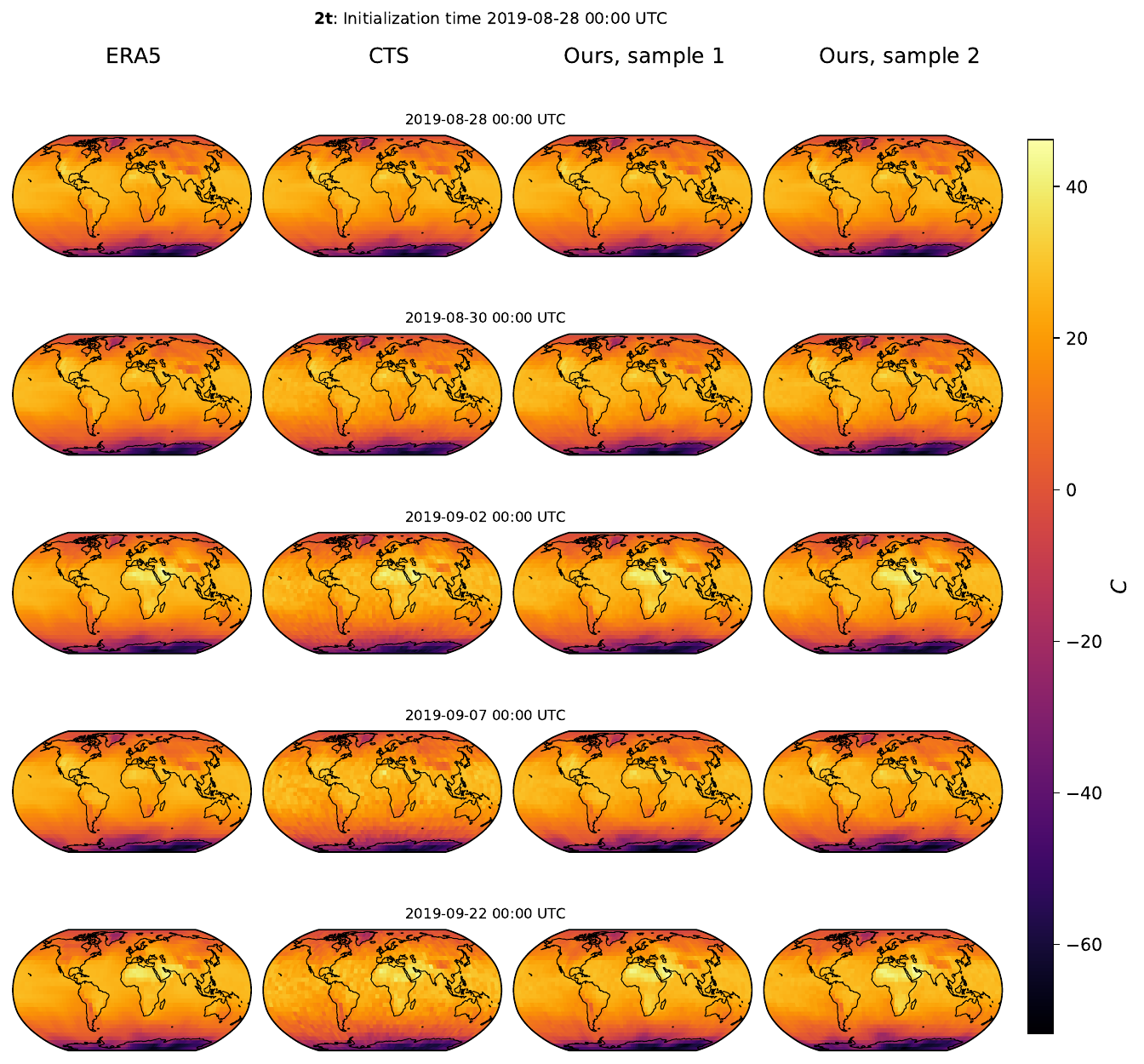}
  \caption{Forecast visualization: 2 metre temperature. Forecast initialized at 2019-08-28 00:00 UTC, with plots corresponding to 2, 5, 10 and 25 day lead times. Two sampled trajectories from our model are shown.} 
  \label{fig:2t}
\end{figure}

\begin{figure}
  \centering
  \hspace*{-1.5cm}
  \includegraphics[height=17cm]{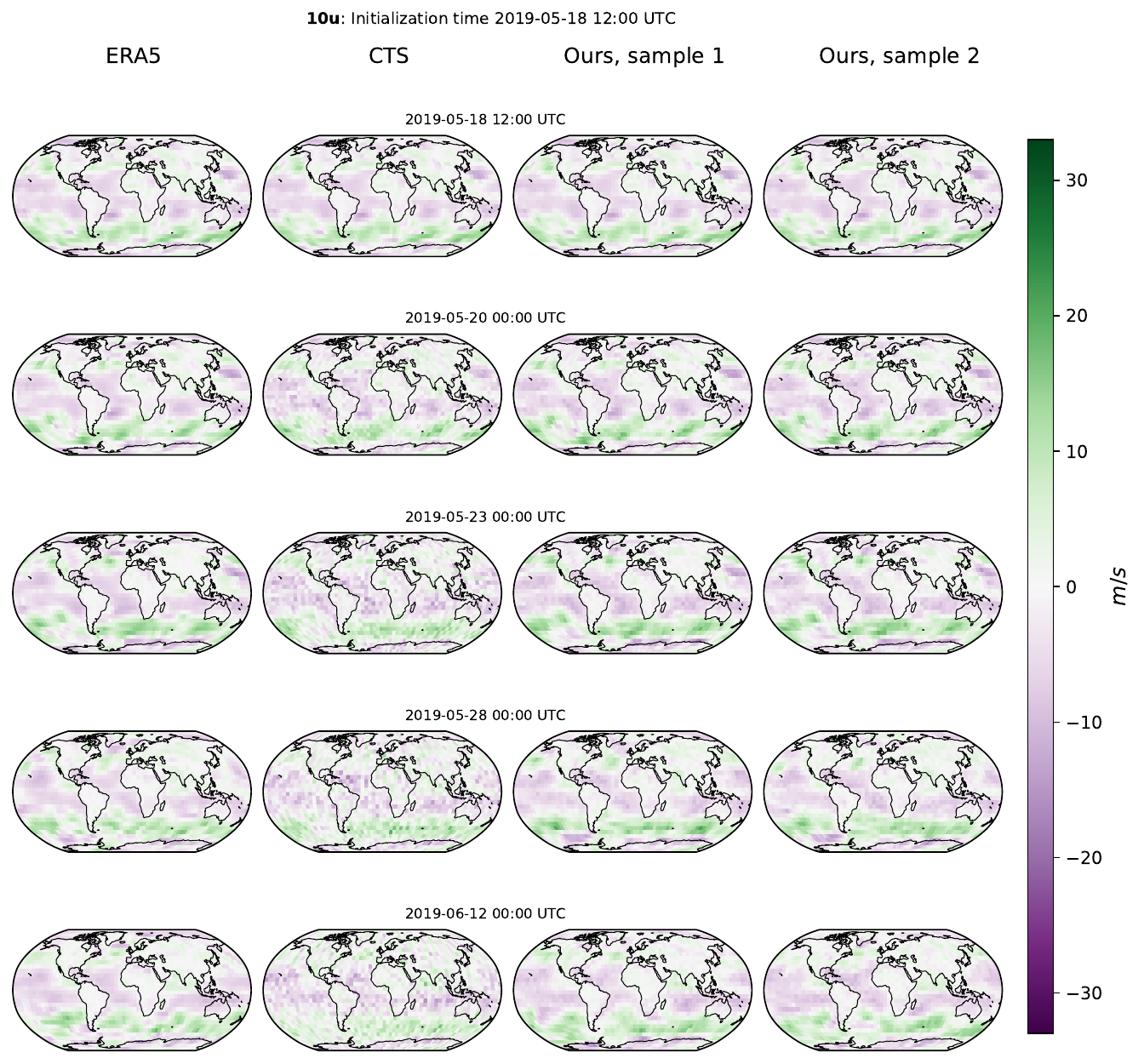}
  \caption{Forecast visualization: 10u. Forecast initialized at 2019-05-18 12:00 UTC, with plots corresponding to 2, 5, 10 and 25 day lead times. Two sampled trajectories from our model are shown.} 
  \label{fig:10u}
\end{figure}

\begin{figure}
  \centering
  \includegraphics[height=5.5cm]{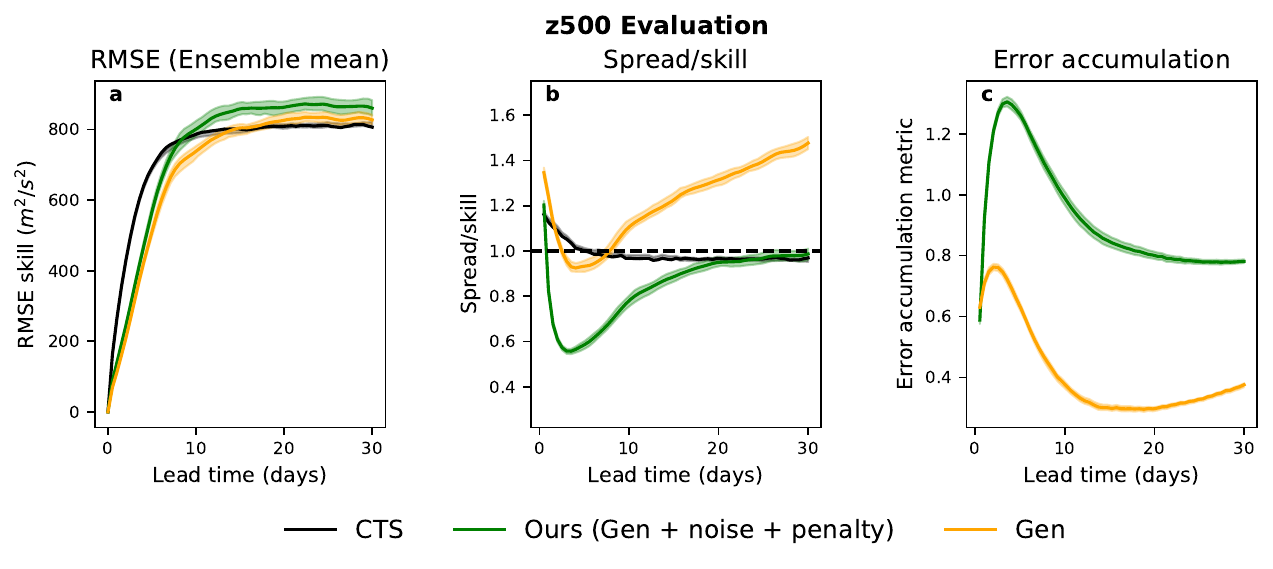}
  \caption{Evaluation of the ensembles produced by a CTS, a generative model and a generative model with our regularization strategy for Z500 (geopotential at pressure level 500) . (a) Ensemble-Mean RMSE skill (lower is better). (b) Spread/skill ratio (closer to 1 is better, lower suggests under-dispersion. (c) Error accumulation metric (lower is better) fails to detect explosion in the generative model. 95\% confidence bands are shown.}
  \label{fig:nwp_z500}
\end{figure}

\begin{figure}
  \centering
  \includegraphics[height=5.5cm]{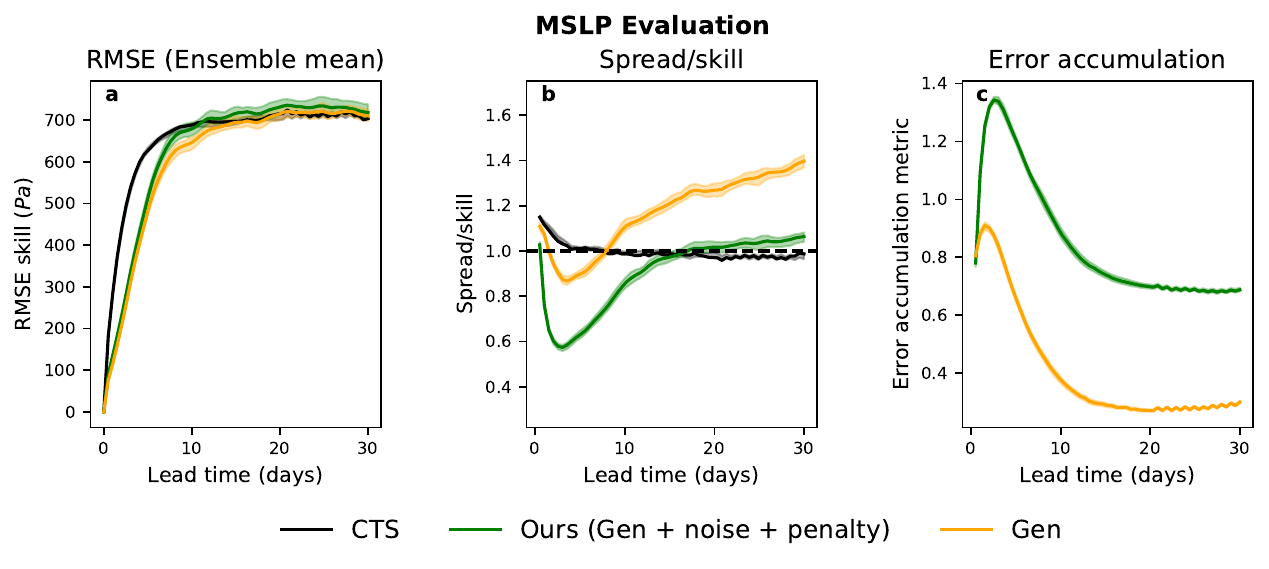}
  \caption{Evaluation of the ensembles produced by a CTS, a generative model and a generative model with our regularization strategy for mean sea-level pressure. (a) Ensemble-Mean RMSE skill (lower is better). (b) Spread/skill ratio (closer to 1 is better, lower suggests under-dispersion. (c) Error accumulation metric (lower is better) fails to detect explosion in the generative model. 95\% confidence bands are shown.}
  \label{fig:nwp_mslp}
\end{figure}

In this section, we provide visualizations for forecasts of T850 (Figure \ref{fig:t850}), z500 (Figure \ref{fig:z500}), 2T (Figure \ref{fig:2t}) and 10u (Figure \ref{fig:10u}). In each case, we show a sample from the CTS, and two sampled trajectories from our model. In Figures \ref{fig:nwp_z500}--\ref{fig:nwp_mslp}, we plot RMSE, spread/skill and our error accumulation metric for z500 and MSLP.

\section{Analysing the error accumulation metric in Figures \ref{fig:nwp_z500}--\ref{fig:nwp_mslp}}
\label{section:diagnostic}

The error accumulation metric for our model is notably worse compared to the base generative model in Figures \ref{fig:nwp_z500}--\ref{fig:nwp_mslp} after lead times of 20 days, despite our model having better RMSE and spread/skill scores. This issue can be traced to deficiencies in the CTS, which are not made apparent in the spread/skill. Since our error accumulation metric measures models against the CTS reference, a poor CTS results in a poor approximation of the desired equation \ref{eq:err_acc_kl}. 

\paragraph{The definition of spread/skill in equation \ref{eq:spread_skill} fails to capture location-specific reliability.} As discussed in \citet{leutbecher2008ensemble}, a reliable forecast will make RMSE errors that match its spread. Therefore, we wish for a forecast's errors at a particular lead time, longitude and latitude, to match its spread there. The standard spread/skill definition in equation \ref{eq:spread_skill} computes compares the RMSE, averaged over latitude and longitude, to the spread, also averaged over latitude and longitude. This averaging masks whether spread and skill are well-matched at each prediction location. 

Spread/skill per location, as defined here, provides more granular detail on location-specific performance:

\begin{equation}
    \textrm{Spread/skill per location} := \frac{1}{M}\sum_m \frac{1}{|G|} \sum_i \sqrt{ \frac{\frac{1}{N} \sum_n (x^n_{i,m} - \bar{x}_{i,m})^2}{(\Bar{x}_{i,m} - x^{O}_{i,m})^2} }
    \label{eq:new_spread_skill}
\end{equation}

The main difference compared to equation \ref{eq:spread_skill} is that the averaging in equation \ref{eq:new_spread_skill} is done after the spread/skill has been computed for each location.

\paragraph{Location-specific metrics highlight the limitations of the CTS.}

\begin{figure}
  \centering
  \includegraphics[height=10cm]{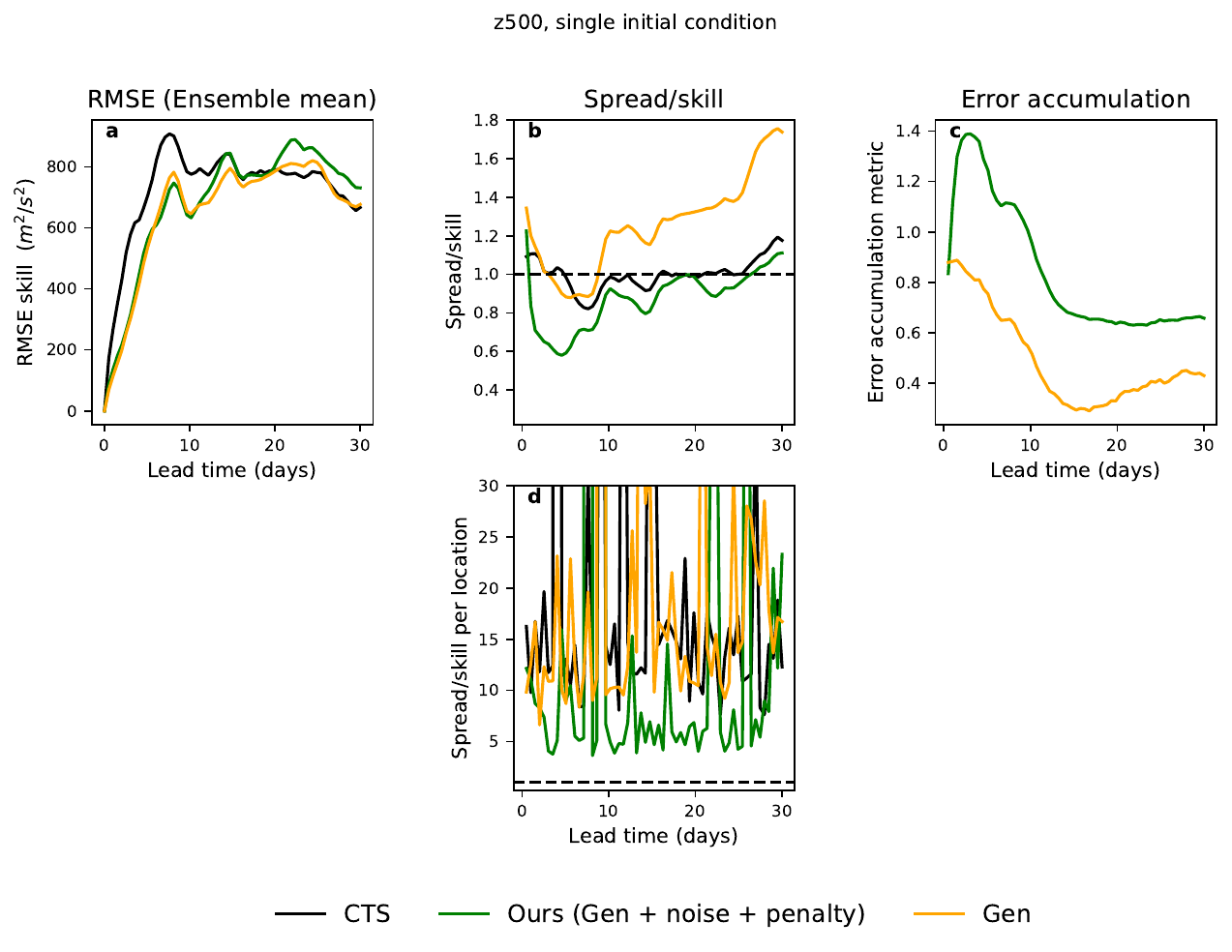}
  \caption{Evaluation of the ensembles produced by a CTS, a generative model and a generative model with our regularization strategy for z500, for a single initial condition from the test set. (a) Ensemble-Mean RMSE skill (lower is better). (b) Spread/skill ratio (closer to 1 is better, lower suggests under-dispersion. (c) Error accumulation metric (lower is better) fails to detect explosion in the generative model. (d) Spread/skill per location makes the differences between the models more evident.}
  \label{fig:spread_skill_location}
\end{figure}

In Figure \ref{fig:spread_skill_location}, we evaluate forecasts made from a single initial condition in the test set. The plot resembles Figure \ref{fig:nwp_z500} in that our model has better spread/skill than the base generative model for lead times greater than 15 days. Both have similar RMSEs, but ours has a worse error accumulation metric (Figure \ref{fig:spread_skill_location}c). In the spread/skill per location (Figure \ref{fig:spread_skill_location}d), our model generally outperforms (lower) both the CTS and the base generative model, with the CTS performing similarly to the generative model.

\begin{figure}
  \centering
  \hspace*{-1.5cm}
  \includegraphics[height=7cm]{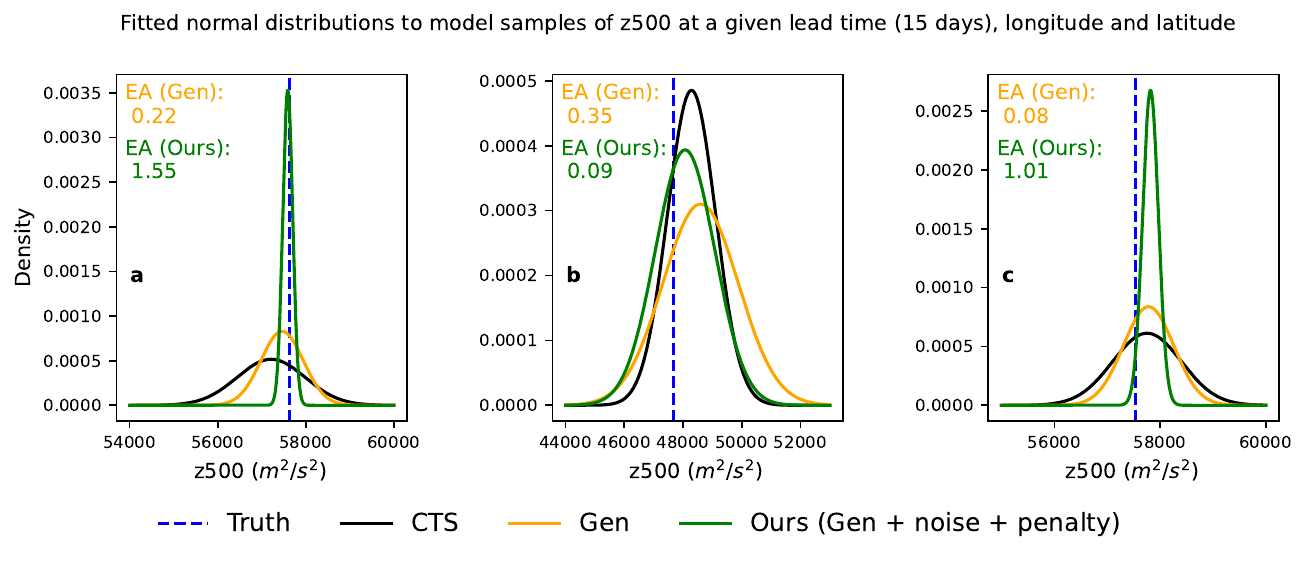}
  \caption{Samples from models for a particular initial condition at a lead time of 15 days, at three sets of longitude and latitude coordinates. Values of the error accumulation metric are shown in the plots. Our model has well-calibrated spread and error in general, but is penalized by the error accumulation metric due to the deficiencies in the CTS.}
  \label{fig:location_samples}
\end{figure}

In Figure \ref{fig:location_samples}, we analyse the distributions from our models for a given initial condition, lead time (15 days), and three sets of longitude and latitude coordinates. We fitted a normal distribution to the samples, which is plotted. In all cases, our model shows well-calibrated error and spread. In Figures \ref{fig:location_samples}a and c, the CTS has deficiencies since it is over-dispersive at these locations. However, because the generative model's distribution is more similar to the CTS, it is the generative model that is assigned a lower error accumulation score.  

\begin{figure}
  \centering
  \includegraphics[height=10cm]{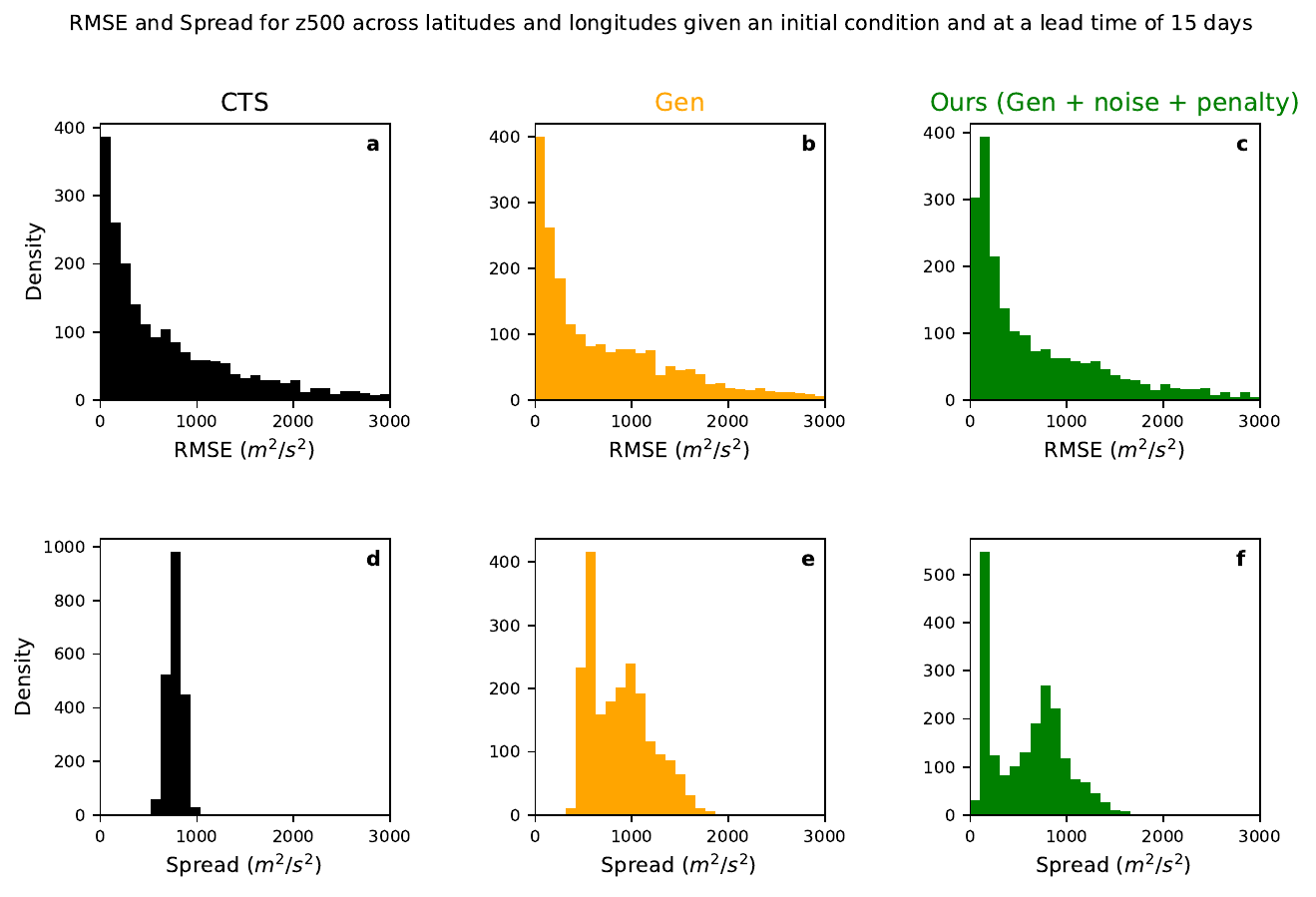}
  \caption{RMSE and spread histograms for z500. Each item composing the histogram corresponds to the RMSE (or spread) for that model at a given latitude and longitude, for the lead time of 15 days and a given initial condition. Ideally, the top and bottom histograms would match. }
  \label{fig:z500_histograms}
\end{figure}

Finally, in Figure \ref{fig:z500_histograms}, we plot histograms of ensemble member mean and standard deviations at each latitude and longitude for a lead time of 15 days, using the same initial condition as in Figure \ref{fig:location_samples}. Ideally, if spread and skill are well-calibrated across locations,  these histograms should look similar. However, this is not the case for any of the models. Notably, only our model has density around small spreads. The CTS model's deficiencies in spread are also evident, indicating areas for improvement. Currently, a single $\sigma$ is learned per variable per lead time for the CTS, meaning the same $\sigma$ is used for all grid cells when modelling z500 at a lead time of 15 days. This contrasts with the mean function, which learns a different mean for each grid cell. The CTS plots suggest a better CTS could be achieved if $\sigma$ was also allowed to vary with latitude and longitude.

\paragraph{The poor CTS acts as a poor reference model in equation \ref{eq:err_acc_kl_approx}.} Equation \ref{eq:err_acc_kl_approx} requires on a good CTS to closely approximate equation \ref{eq:err_acc_kl}. As noted above, the CTS is inadequate. The reason the generative model shows a lower error accumulation score is because its samples are closer to those of the CTS. Improving the CTS would result in the error accumulation metric reflecting the desired detection of explosions in the generative model, and our model's better performance at longer lead times. Moreover, a better CTS should allow for a more useful regularization penalty, which may further improve results.